\documentclass[smallextended]{svjour3}      
\usepackage{amssymb}
\usepackage{amsmath}
\usepackage{pdflscape}
\usepackage{array,multirow}
\usepackage{caption}
\usepackage{rotating}
\usepackage{graphicx}
\usepackage{subfig}
\usepackage{tabularx}
\usepackage{multirow}
\usepackage{lmodern}%
\usepackage{changepage}

\smartqed 
\usepackage{graphicx}

\begin{document}
	
	\title{ A new asymmetric $\epsilon$-insensitive pinball loss function based support vector quantile regression model}

	\titlerunning{~~~~}      
	
	\author{ Pritam Anand \and Reshma Rastogi (nee Khemchandani) \and Suresh Chandra. 
	}
	
	\institute{   Pritam Anand \\
		Research Scholar, \at
		Faculty of Mathematics and Computer Science,\\
		South Asian University, New Delhi-110021.\\
		\email{ltpritamanand@gmail.com ~.} 
		\and 
		R. Rastogi (nee Khemchandani)   \\
		Assistant Professor,             \at
		Faculty of Mathematics and Computer Science,\\
		South Asian University, New Delhi-110021. \\
		\email{reshma.khemchandani@sau.ac.in .}  \\
		Tel No. 01124195147        
		\and
		Suresh Chandra\at
		Ex-Faculty, Department of Mathematics\\
		Indian Institute of Technology Delhi.
		New Delhi-110016.\\
		\email{chandras@maths.iitd.ac.in}                            
	}
	
	\date{Received: date / Accepted: date}

	\maketitle

\begin{abstract}
In this paper, we propose a novel asymmetric $\epsilon$-insensitive pinball loss function for quantile estimation.  There exists some pinball loss functions which attempt to incorporate the $\epsilon$-insensitive zone approach in it but, they fail to extend the $\epsilon$-insensitive approach for quantile estimation in true sense.  The proposed asymmetric $\epsilon$-insensitive pinball loss function can make an asymmetric $\epsilon$- insensitive zone of fixed width around the data and divide it using $\tau$ value for the estimation of the $\tau$th quantile. The use of the proposed asymmetric $\epsilon$-insensitive pinball loss function in Support Vector Quantile Regression (SVQR) model improves its prediction ability significantly. It also brings the sparsity back in SVQR model.
 Further,  the numerical results obtained by several experiments carried on artificial and real world datasets empirically show the efficacy of the proposed  `$\epsilon$-Support Vector Quantile Regression' ($\epsilon$-SVQR) model over other existing SVQR models. 

\end{abstract}

\keywords{\texttt { Quantile Regression},  pinball loss function , Support Vector Machine, $\epsilon$-insensitive loss function. }

\section{Introduction}
Given  training set $T= \{ (x_i,y_i): x_i \in \mathbb{R}^n, y_i \in \mathbb{R},~ i=1,2...,l~ \}$, the problem of regression is concerned with finding a function $f(x)$ which estimates the conditional mean of $y$ given $x$. But, only the estimation of the conditional mean function is not enough to give a full description about the stochastic relationship between the target and response variables. Therefore, in many applications, we are interested in the estimation of the conditional quantile functions $f_{\tau}(x)$ as well. 

       The quantile regression problem had intialy been studied in 1978 by Koenkar and Bassett\cite{quantile1}, which was later detalied and discussed in (Koenker, \cite{quantile2}).   
       Koenkar and Bassett \cite{quantile1} proposed the use of pinball loss function for the estimation of the conditional quantile function  $f_{\tau}(x)$. The pinball loss function is an asymmetric loss function which, for a given quantile $\tau \in (0,1)$, is defined as
       \begin{equation}
       P_{\tau}(u) ~=~ \begin{cases}
       \tau u ~~~~~~~~~~\mbox{if}~~ u > 0, \\
       (\tau-1)u~~~ \mbox{otherwise}.
       \end{cases}
       \label{pinballloss}
       \end{equation}
       
        But, Takeuchi et al \cite{quantile3} were first to initiate the study of the quantile regression problem in a non-parametric framework\cite{quantile3}. It also establishes that a minimizer of  pinball loss function (\ref{pinballloss}) asymptotically converges to the real quantile function under a very general conditions.  Their formulation consist of  minimization of the pinball loss function (\ref{pinballloss}) along with a regularization term  for the estimation of the conditional quantile function $f_{\tau}(x)$ . Like Support Vector Regression (SVR) model (Vapnik et al.,\cite{svr1})(Drucker et al.,\cite{svr2}),(Gunn, \cite{GUNNSVM}) their proposed Support  Vector Quantile Regression (SVQR) model is also consistent with the Structural Risk Minimization (SRM) principle (Vapnik, \cite{statistical_learning_theory}).
        
          It is well known that sparsity is a very desirable property in a regression model. A sparse regression model uses few training data points for the construction of the regression function and is very  time efficient in the prediction of the responses of test data points. Unlike $\epsilon$-SVR model, the SVQR model lacks sparsity as all of the training data points contribute to the empirical risk in the pinball loss function. That is why, a SVQR model, which can use the $\epsilon$-insensitive approach efficiently is required for increasing its generalization ability and bringing  the sparsity back in the model.

                        \begin{figure}
                        	\centering
                            \subfloat[] {\includegraphics[width=2.5in,height=1.65in]{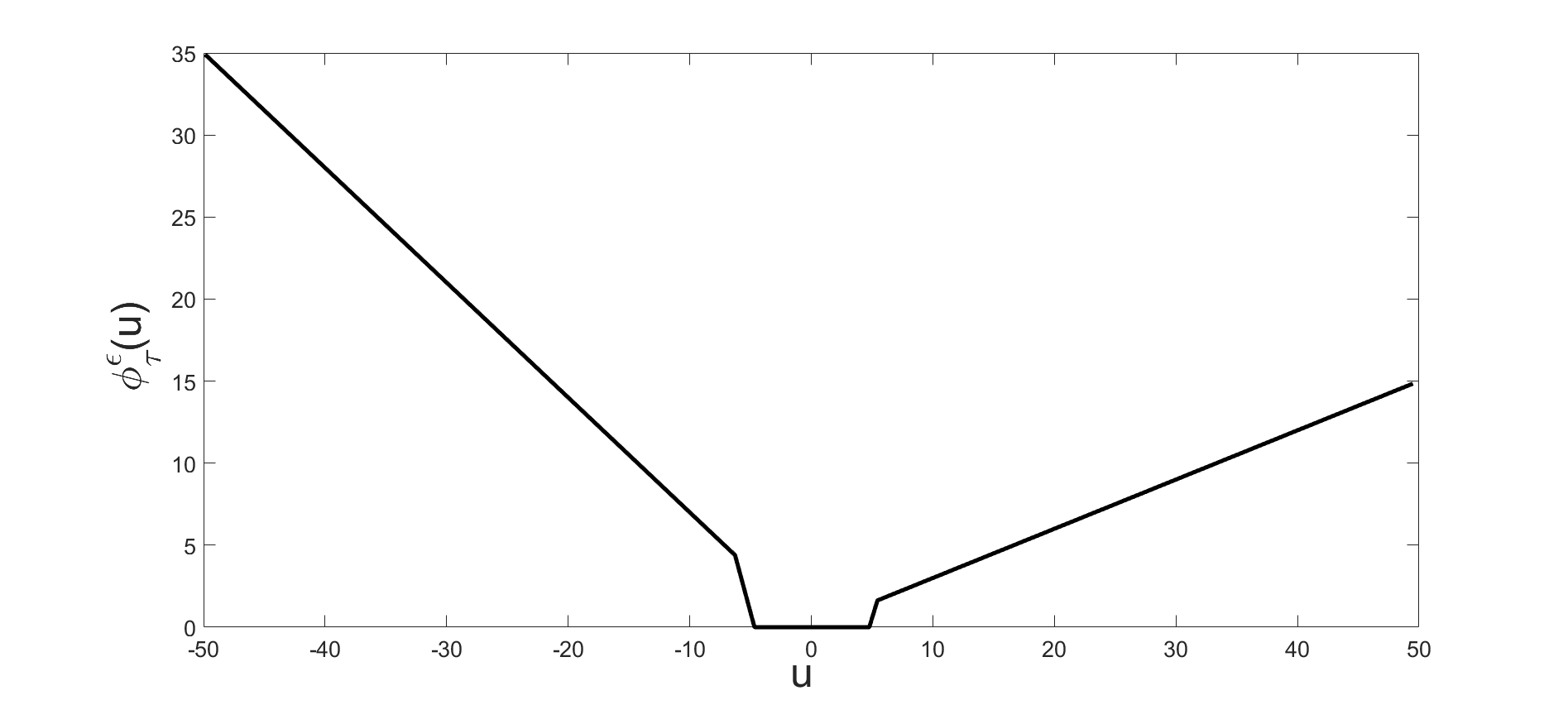}}
                                                    	\subfloat[] {\includegraphics[width=2.5in,height=1.65in]{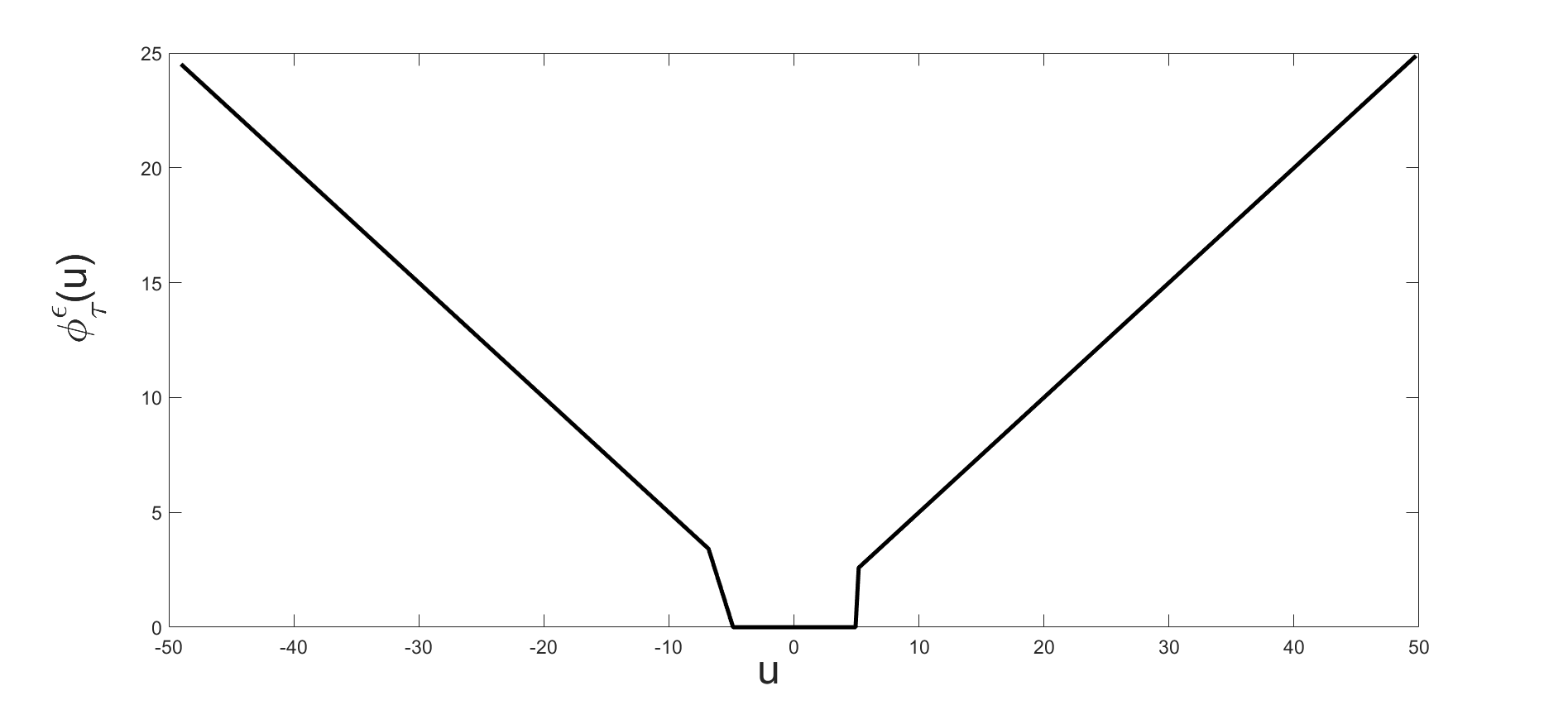}}
                            	\caption{Symmetric $\epsilon$-insensitive pinball loss function described in (Takeuchi and  Furuhashi, \cite{noncrossqsvr}) for (a) $\tau =0.3$ (b) $\tau= 0.5$ with $\epsilon$=5.}
                                \label{pinlossnoncross}
                            \end{figure}

                         The idea of using of  $\epsilon$-insensitive tube in the SVQR model seems to be obvious and has been described at number of places in the literature. But, we have not found any formulation which extends the idea of the $\epsilon$-insensitive tube in SVQR model in  its true sense.
                        
                         Takeuchi and  Furuhashi considered the $\epsilon$-insensitive pinball loss function for estimation of the non-crossing quantile in their work (Takeuchi and  Furuhashi, \cite{noncrossqsvr}).They combined the symmetric $\epsilon$-insensitive tube with the asymmetric pinball loss function by considering the following loss function
                         \begin{equation}
                        \phi_{\tau}^{\epsilon}(u)= 
                        \begin{cases}
                        (1-\tau)|u|, if~~u < -\epsilon \\
                        0,         ~~~~~~~~~~if~~ |u| \leq \epsilon \\
                        \tau|u|,~~~~~if ~~~~u> \epsilon
                        \end{cases}
                        \end{equation}
                        for estimation of the non-crossing quantiles function. However, they had also admitted there that the introduction of  $\epsilon$-tube is unfavorable for  estimation of the conditional quantile estimator.  One of the possible reason for this could be the symmetry of the $\epsilon$-tube i,e. the $\epsilon$-tube is symmetric around the estimated function. We have also plotted the $\epsilon$-insensitive pinball loss function described in (Takeuchi and  Furuhashi, \cite{noncrossqsvr}) in the Figure (\ref{pinlossnoncross}) and found that the proposed loss function is not convex and hence cannot be properly minimized using any convex program. Further, the given loss function doesn't reduce to the Vapnik $\epsilon$-insensitive loss function for $\tau=0.5$. Hu et al, had also considered the similar kind of $\epsilon$-insensitive pinball loss function in their work (Hu et al, \cite{onlinesvqr}) for estimation of quantiles  but here also, the $\epsilon$-tube was symmetric around the estimated function.
                        
                        Seok et al.\cite{sparsequantile}  also attempted to extend the idea of $\epsilon$-insensitive approach in SVQR model.  But for this, they proposed an asymmetric e-insensitive pinball loss function for quantile estimation which is as follows.
                        \begin{equation}
                        h_{\tau}(u)= 
                        \begin{cases}
                        0, ~~~~~~~~~~~~~~~~~~~if~~~ \frac{\tau}{\tau-1}\epsilon \leq u \leq \frac{1-\tau}{\tau}\epsilon.\\
                        \tau u -(1-\tau)\epsilon, ~~~~if~~~u \geq \frac{1-\tau}{\tau}\epsilon.  \\
                        (\tau-1)u-\tau\epsilon,~~~~if ~~~~u\leq \frac{\tau}{\tau-1}\epsilon.
                        \end{cases}
                        \end{equation}
                         Their resulting formulation was termed with 'Sparse Support Vector Quantile Regression' (Sparse SVQR) model.  The Sparse SVQR model was able to obtain sparse solution. The e-insensitive pinball loss function of Seok et al.\cite{sparsequantile} can make an asymmetric $\epsilon$-insensitive zone around the estimated function.  But, there is still major problem in it.  The width of the $\epsilon$-insensitive zone in e-insensitive pinball loss function of Seok et al.\cite{sparsequantile} varies with the $\tau$ values where as it should ideally vary with the variance present in the response values of the training data. It also makes the selection of the good $\epsilon$-value difficult in practice. Further, the Sparse SVQR model requires the tuning of different choices of $\epsilon$ for the prediction of different conditional quantile function for a given training set. 
                           
                          Park and Kim \cite{quantilerkhs} had also proposed a similar kind of improvement in the $\epsilon$-insensitive pinball loss function for quantile regression model in reproducing kernel Hilbert space. They have proposed the following loss function
                           \begin{eqnarray}
                           \rho_{\tau}^{\epsilon}(u)=  max(0, P_{\tau}(u)) 
                         ~~ = \begin{cases}
                           P_{\tau}(u)-\epsilon, ~~~if~~P_{\tau}(u) > \epsilon. \\
                           0,   ~~~~~~~~~~~~~~~    otherwise. \\
                           \end{cases}
                           \end{eqnarray}
                           
                          Similar to the e-insensitive pinball loss function of Seok et al.\cite{sparsequantile}, the width of the  $\epsilon$-insensitive zone in the loss function of  Park and Kim \cite{quantilerkhs} also varies with the $\tau$ values.

                      We have realized the need of developing an $\epsilon$-insensitive pinball loss function  which can extend the $\epsilon$-insensitive approach in pinball loss function in true sense for the quantile estimation. For this, we have proposed a novel asymmetric $\epsilon$-insensitive pinball loss function to be used for quantile estimation in this paper. For a given $\tau \in (0,1)$ , the proposed asymmetric $\epsilon$-insensitive pinball loss function is given by
                      \begin{equation}
                      L_{\tau}^{\epsilon}(u)= max(~-(1-\tau)(u+\tau\epsilon),~0~,~\tau(u-(1-\tau)\epsilon)~)
                      \end{equation}
                      For the problem of the quantile regression and given quantile $\tau \in (0,1)$,  it can be better understood in the following form.
                      \begin{equation}
                      L_{\tau}^{\epsilon}(y_i,x_i,w,b)= 
                      \begin{cases}
                      -(1-\tau)(y_i-(w^Tx_i+b)+ \tau\epsilon), ~if~~y_i-(w^Tx_i+b) < -\tau\epsilon. \\
                      0,         ~~~~~~~~~~~~~~~~~~~~~~~~if~~ -\tau\epsilon \leq y_i-(w^Tx_i+b)\leq (1-\tau)\epsilon. \\
                      \tau(y_i-(w^Tx_i+b)-(1-\tau)\epsilon),~if ~y_i-(w^Tx_i+b) >(1-\tau)\epsilon.
                      \end{cases}
                      \end{equation}

                      Unlike other $\epsilon$-insensitive loss functions, the overall width of the $\epsilon$-zone in the proposed asymmetric $\epsilon$-insensitive pinball loss function does not vary for different values of $\tau$ rather, the  division of the  $\epsilon$-insensitive zone along the regressor is dependent on the specific $\tau$ value. The expected  number of training points lying above and below the estimated function decides the length of the $\epsilon$-insensitive zone assigned to below and above the estimated function.  In this way, the proposed asymmetric $\epsilon$-insensitive pinball loss function  incorporates the concept of the $\epsilon$-insensitive zone in existing SVQR model in true sense.

                      \begin{figure}
                      	\centering
                      	\subfloat[] {\includegraphics[width=3.0in,height=1.65in]{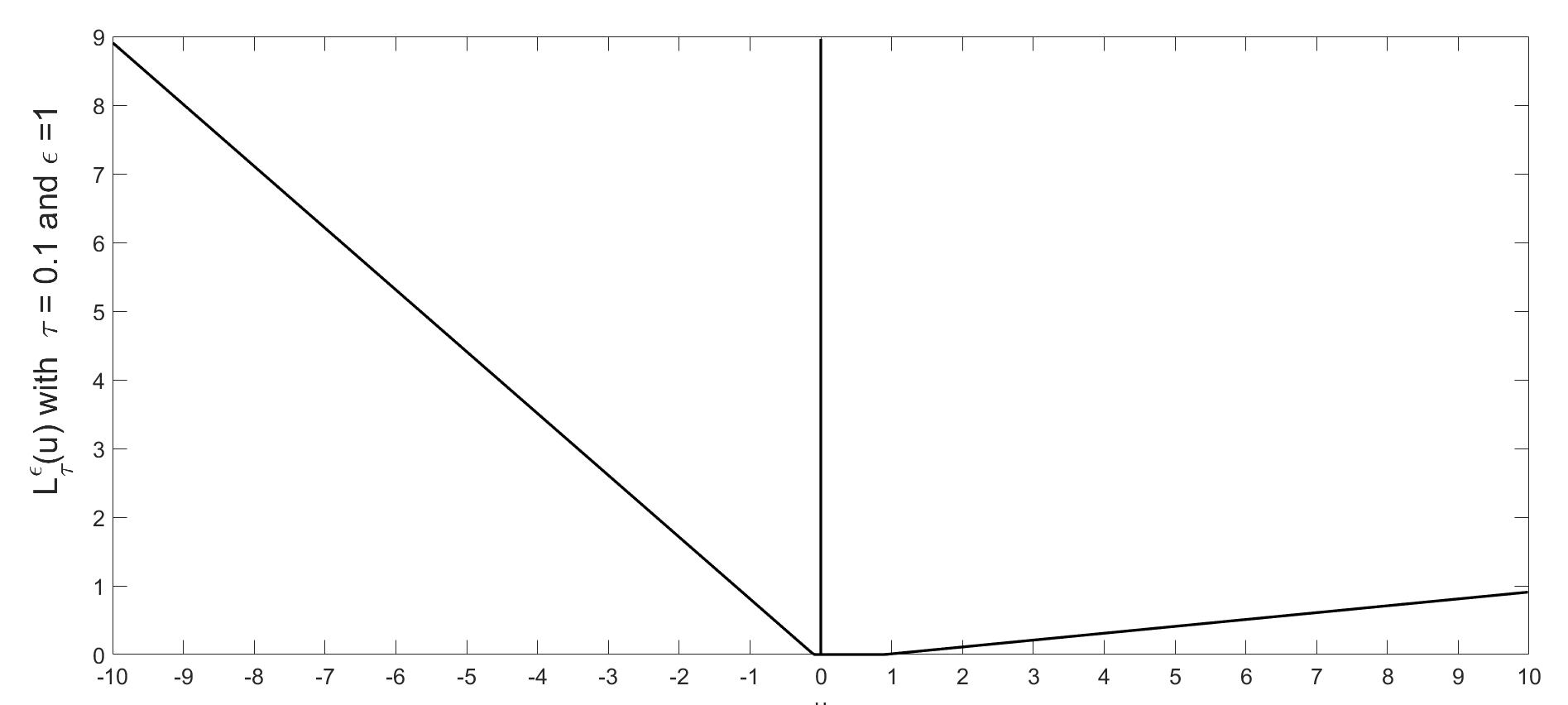}}
                      	\subfloat[] {\includegraphics[width=0.65\linewidth]{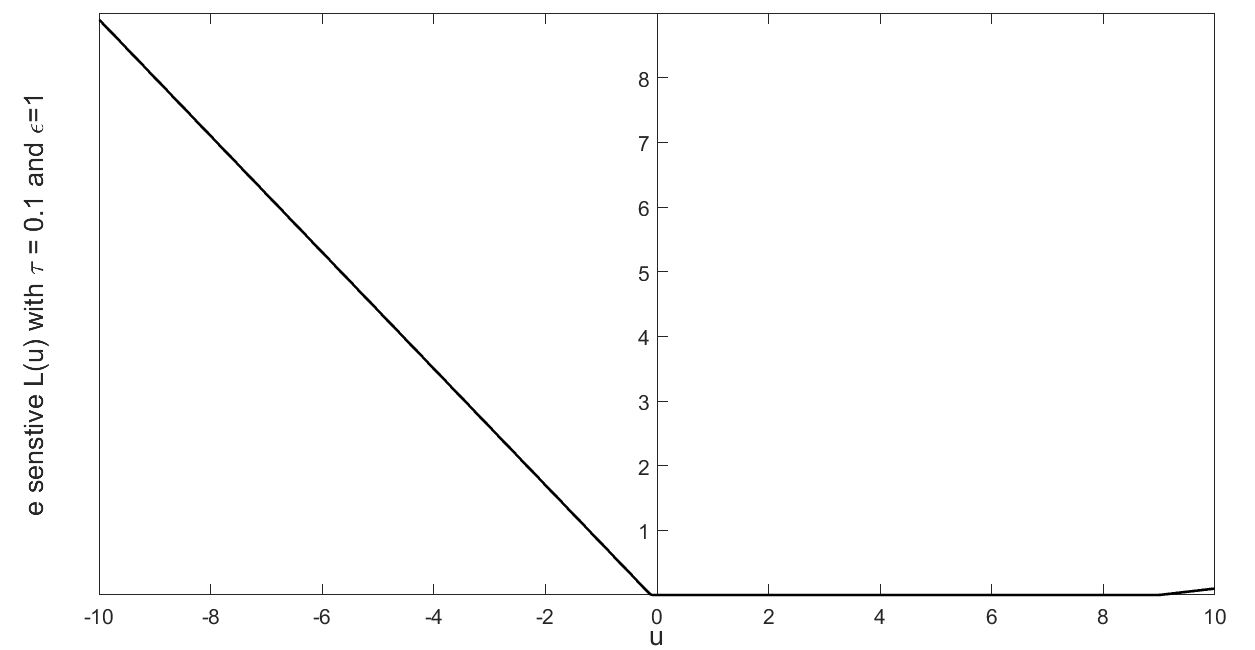}}\\
                      	\subfloat[] {\includegraphics[width=0.65\linewidth]{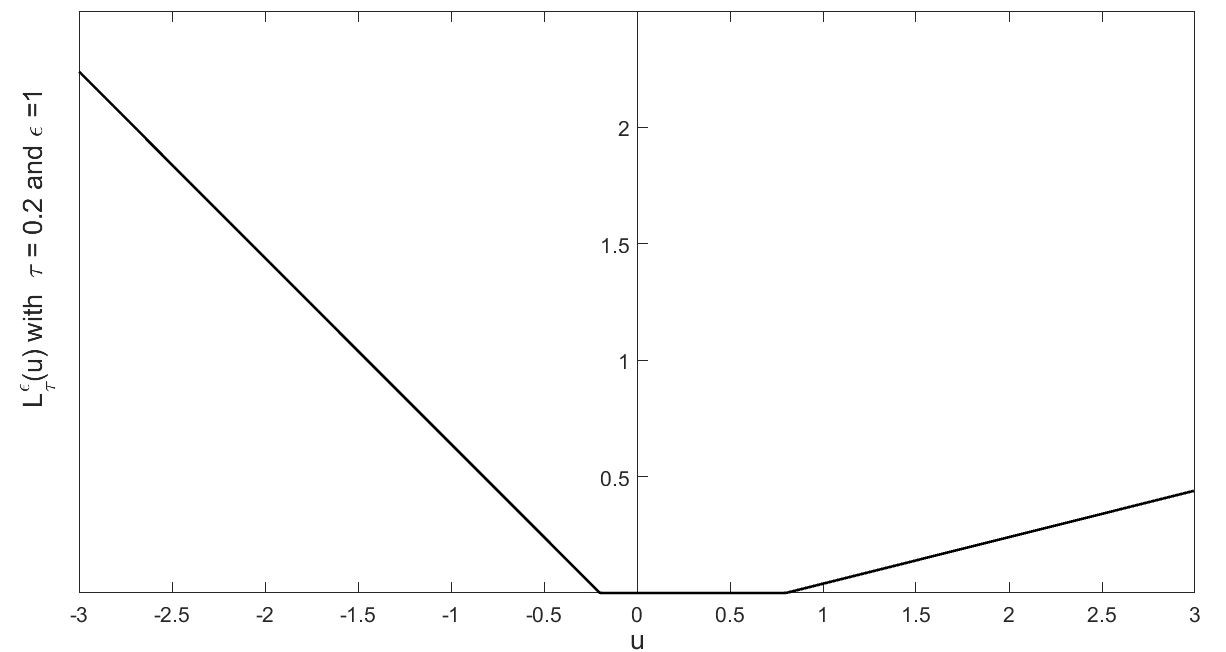}}
                      	\subfloat[] {\includegraphics[width=0.65\linewidth]{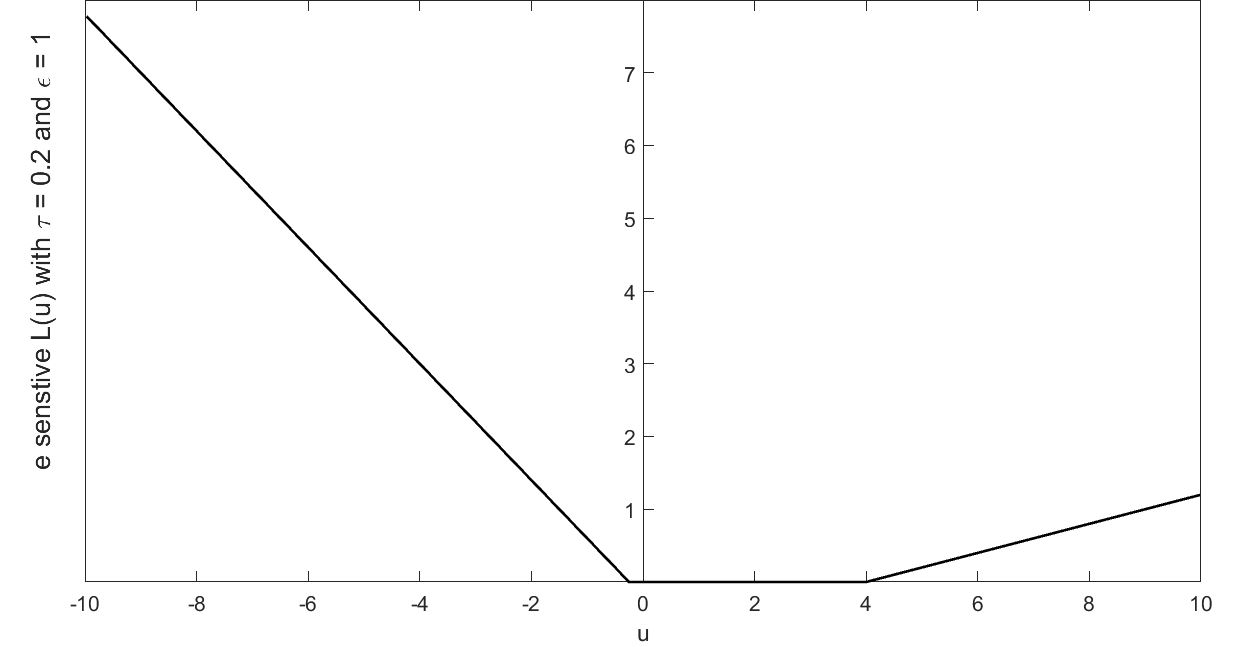}}\\
                      	\subfloat[] {\includegraphics[width=3.0in,height=1.75in]{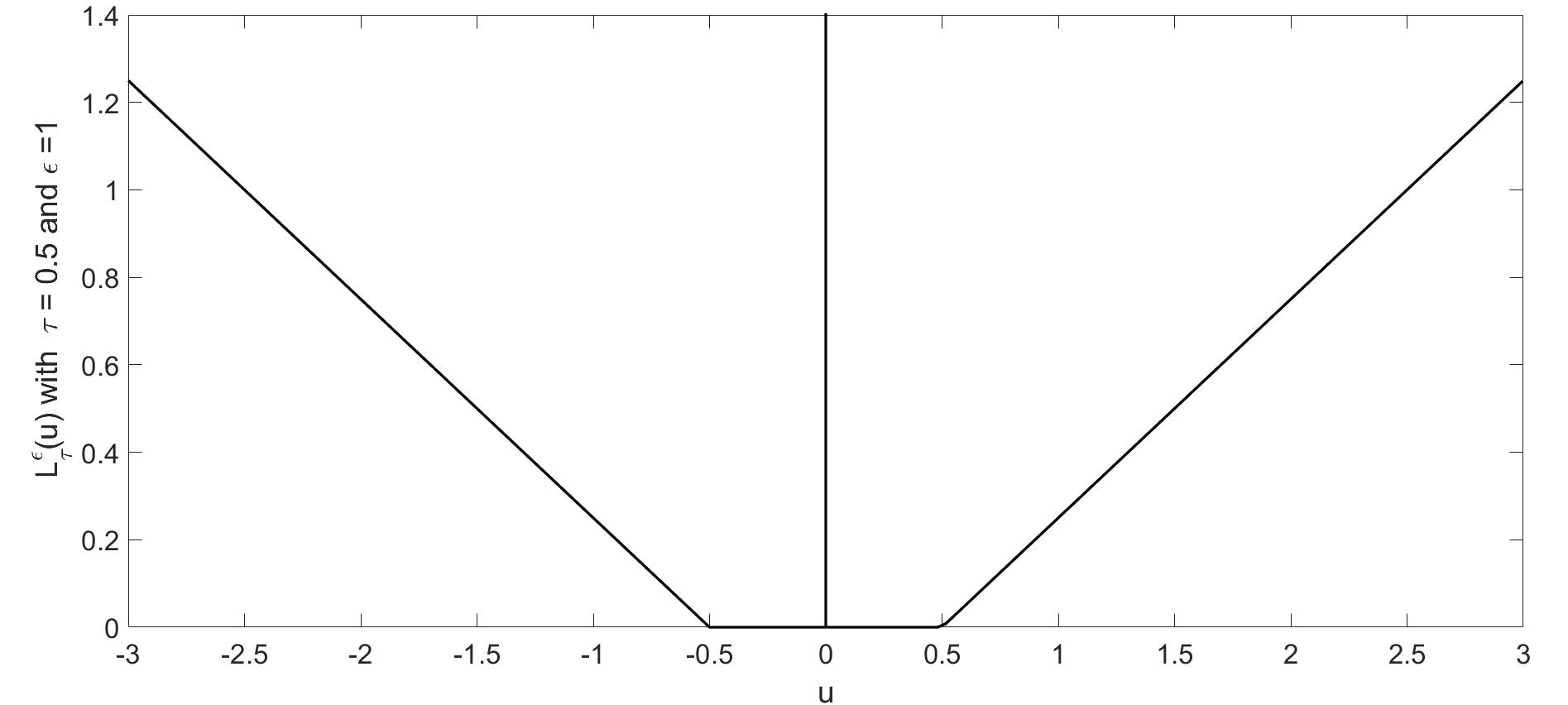}}
                      	\subfloat[] {\includegraphics[width=0.65\linewidth]{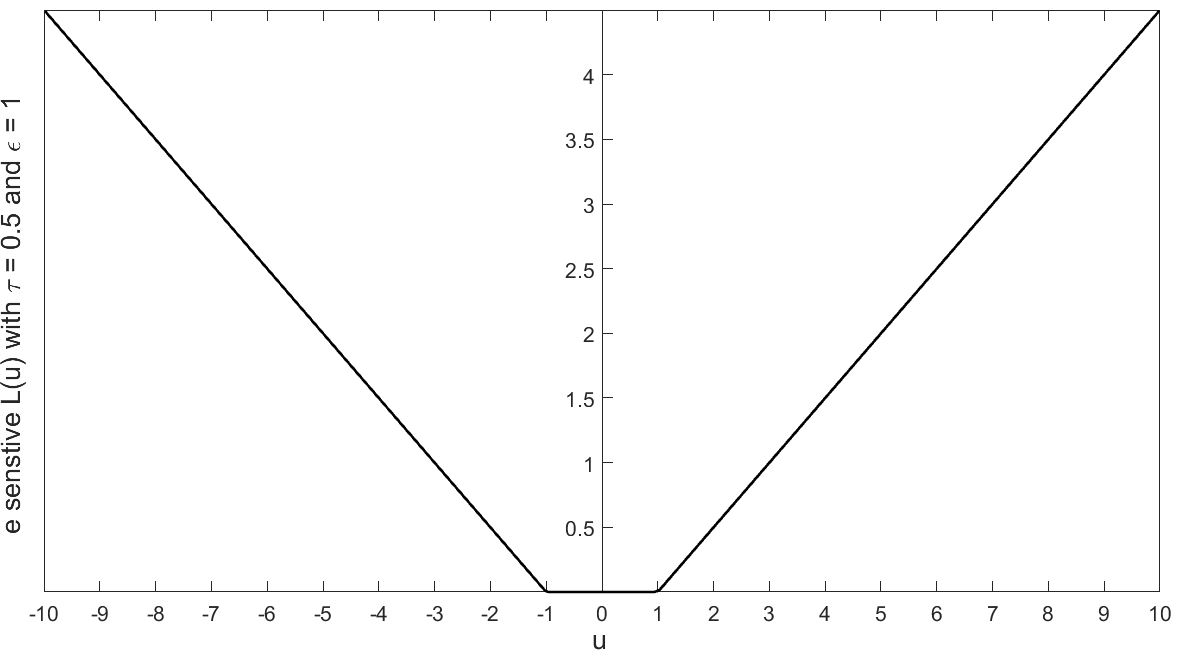}}\\
                      	\subfloat[] {\includegraphics[width=0.65\linewidth]{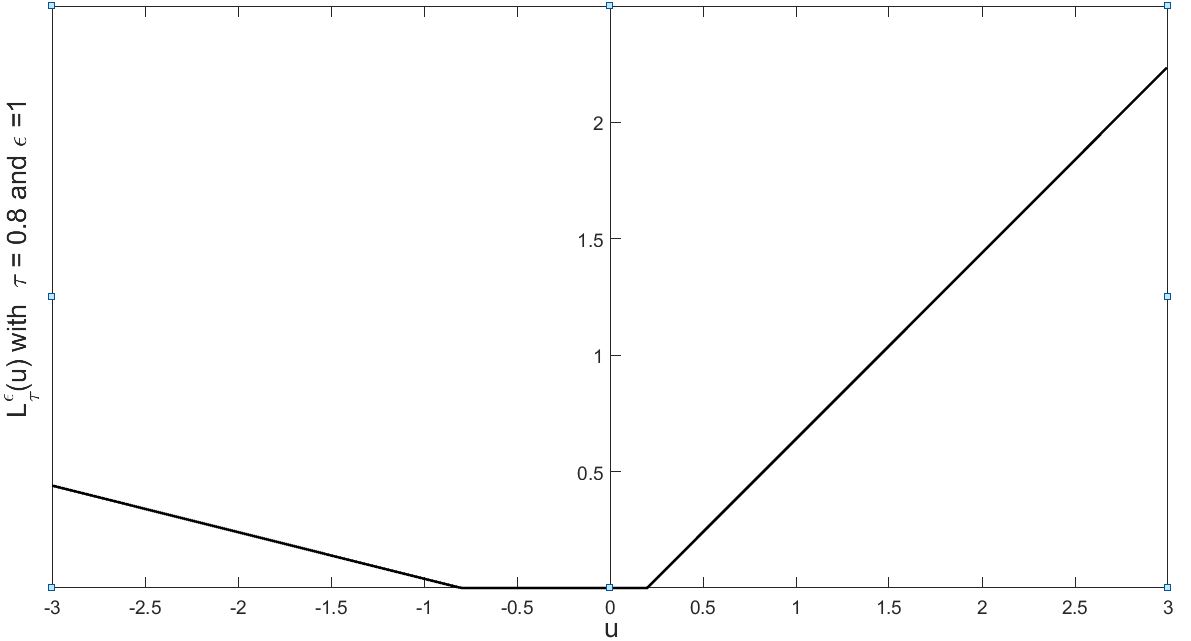}}
                      	\subfloat[] {\includegraphics[width=0.65\linewidth]{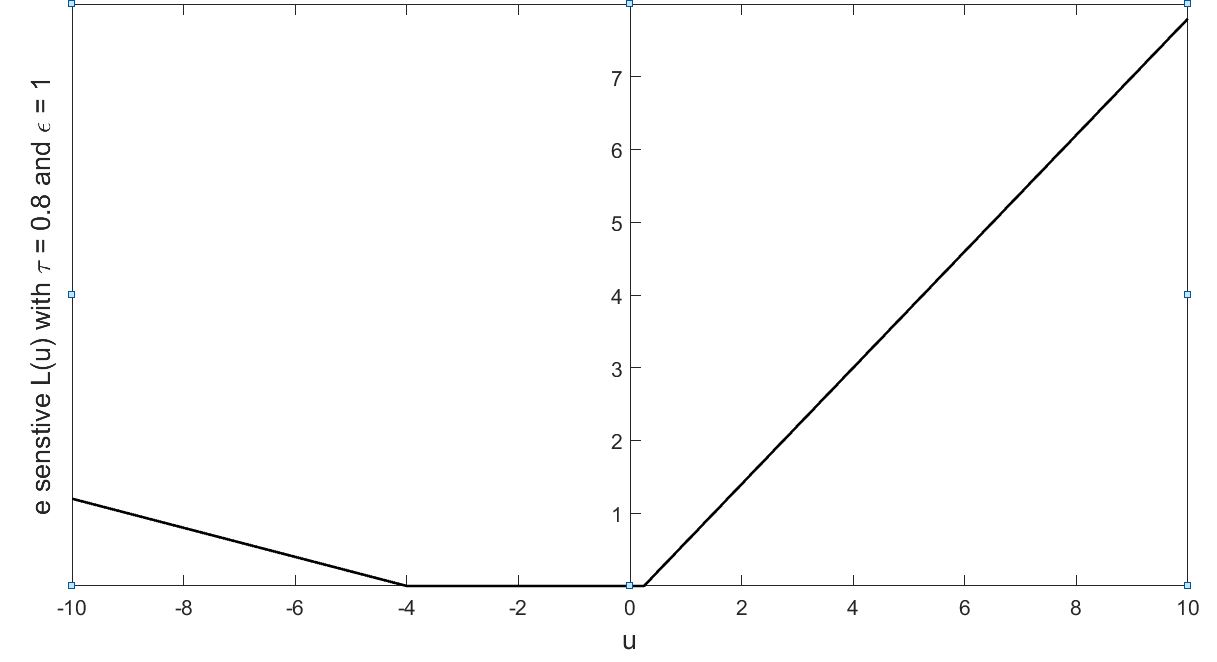}}
                      	\caption{ Comparison of the proposed asymmetric $\epsilon$-pinball loss function (left) and e-insensitive pinball loss function of Seok et al.\cite{sparsequantile} (right)  for (a) $\tau =0.1$ (b) $\tau= 0.2$ (c) $\tau = 0.5$  and (d) $\tau = 0.8$  with fixed $\epsilon$=1.}
                      	\label{ourloss}r
                      \end{figure}
                      
                       Figure \ref{ourloss} shows the comparison of the proposed asymmetric $\epsilon$-pinball loss function and existing e-insensitive loss function of Seok et al.\cite{sparsequantile} for different values of $\tau$ with the fixed values of $\epsilon=1$.  From this figure, it can be observed that, unlike the e-insensitive loss function, the total width of the $\epsilon$-insensitive zone is fixed in  the proposed $\epsilon$-pinball loss function in all cases. However, the division of the $\epsilon$-insensitive zone is not symmetric and depends on the specific $\tau$ value chosen. The underlying logic behind this division of the $\epsilon$-insensitive zone is that it should be based on the expected  number of training points lying above and below the estimated regressor. Further, it can be observed that the total width of the $\epsilon$-insensitive zone is not fixed to 1 in the existing e insensitive loss function and  does depend on the $\tau$ value which may lead to the inaccurate result. For example for $\tau=0.1$, the e insensitive loss function assigns very large insensitive zone towards the upside of the estimated regressor which makes it to ignore most of the training points lying up side  of the estimated regressor and distort the generalization ability of the estimated regressor. This seems to be a major drawback of the e-insensitive loss function and the resulting Sparse SVQR model\cite{sparsequantile}, which has been very well handled in our proposed $\epsilon$-SVQR model.
                        
                         \begin{figure}
                        	\centering
                        	\subfloat[] {\includegraphics[width=2.5in,height=1.65in]{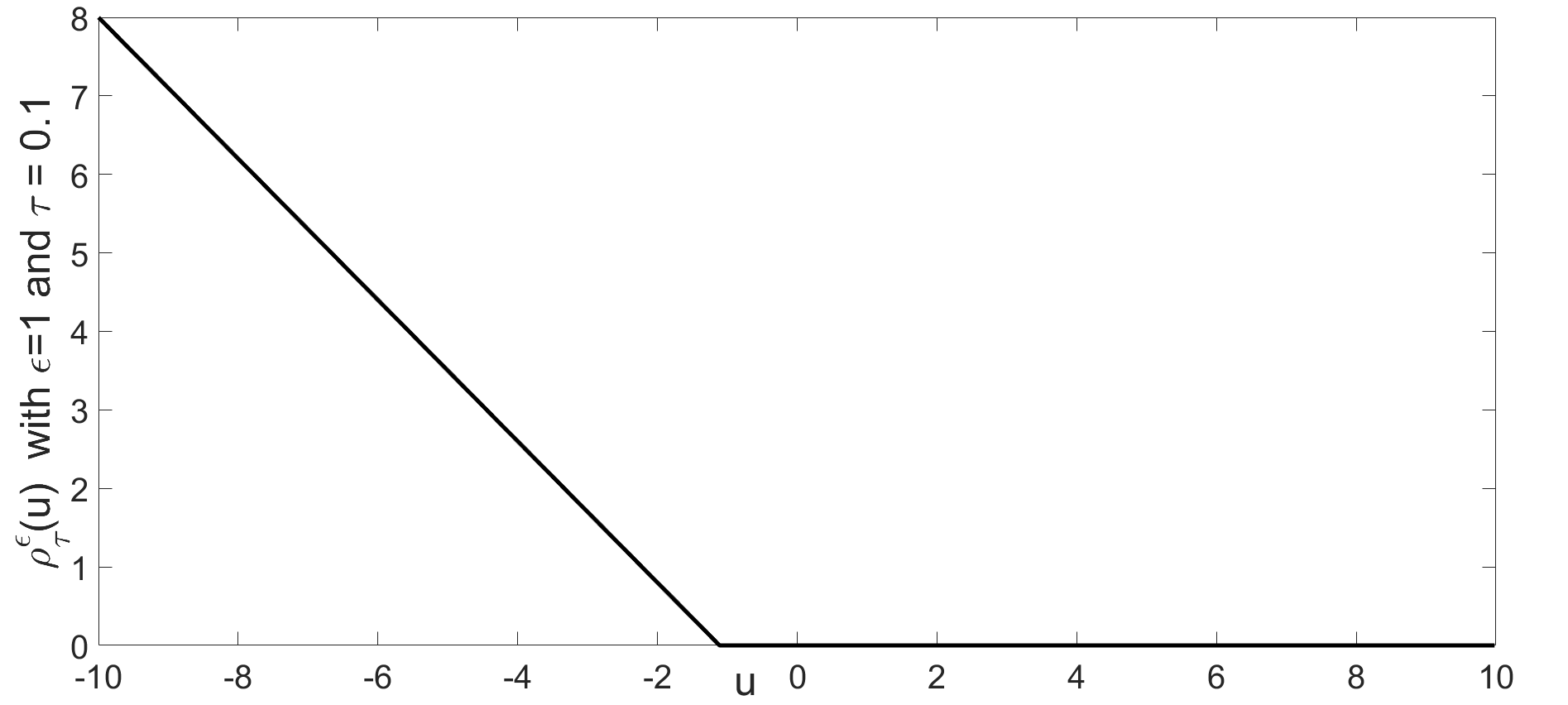}}
                        	\subfloat[] {\includegraphics[width=2.5in,height=1.65in]{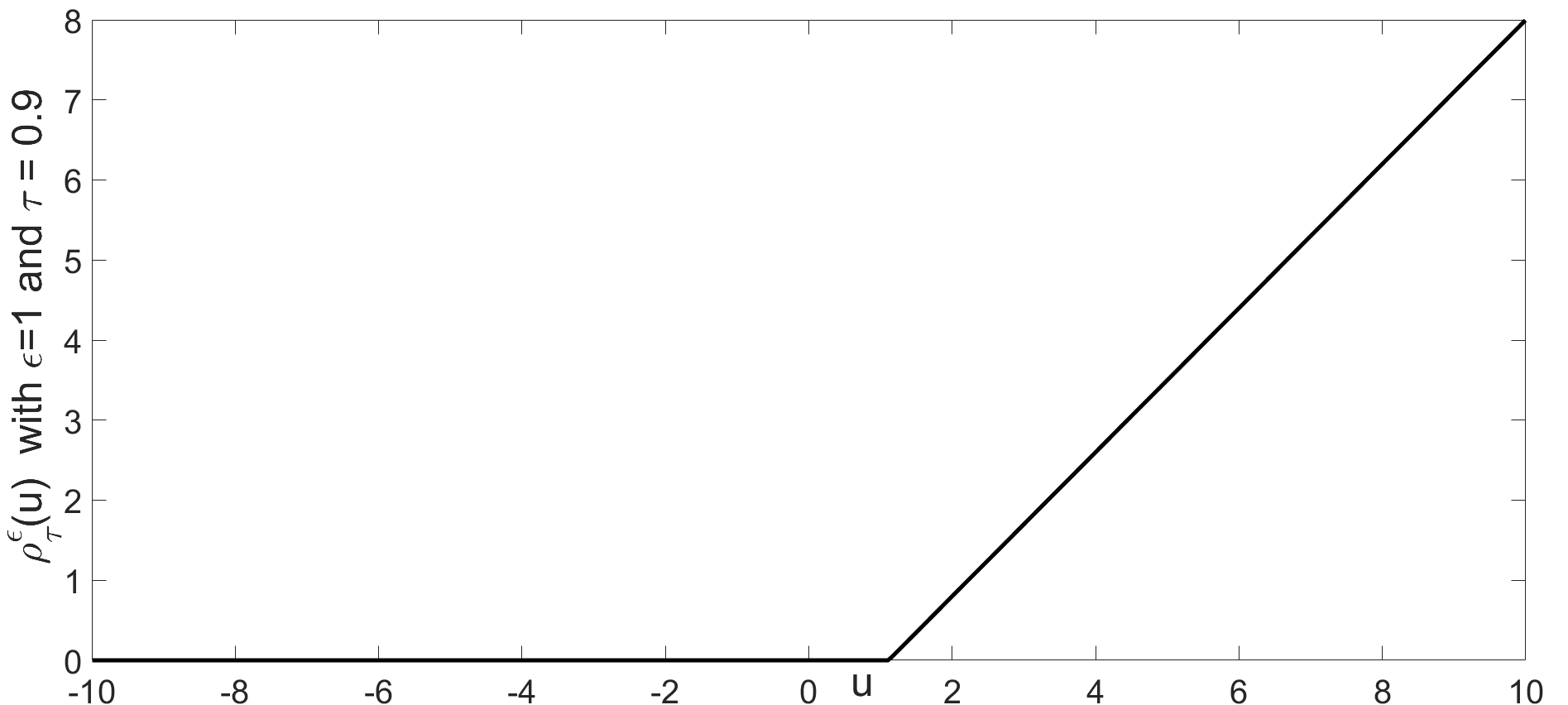}}\\
                        	\caption{ The $\epsilon$-pinball loss function proposed by Park and Kim \cite{quantilerkhs}  for (a) $\tau =0.1$ (b) $\tau= 0.9$ with fixed $\epsilon$=1.}
                        	\label{Parkloss}
                        \end{figure}
                        Figure \ref{Parkloss} shows the plot of $\epsilon$-pinball loss function proposed by Park and Kim \cite{quantilerkhs}  for  $\tau =0.1$ and 0.9 with fixed $\epsilon=1$. Like e-insensitive pinball loss function of Seok et al.\cite{sparsequantile}, the  overall width of the  $\epsilon$-insensitive zone in the loss function of  Park and Kim \cite{quantilerkhs} also varies with the $\tau$ values. It makes the loss function of  Park and Kim \cite{quantilerkhs} not suitable for practice.
                        
                        Further, we  minimize the proposed  asymmetric $\epsilon$-insensitive pinball loss function loss function along with a regularization term in a regression model  for quantile estimation.  We term the resultant regression model with `$\epsilon$-Support Vector Quantile Regression' ($\epsilon$-SVQR) model. The proposed $\epsilon$-SVQR model extends the $\epsilon$-insensitive approach in SVQR model in true sense. The proposed $\epsilon$-SVQR model\cite{sparsequantile}  considers an asymmetric $\epsilon$-insensitive zone around the  quantile regressor and ignores  data points which lie in this zone. The data points which lie outside of the $\epsilon$-insensitive zone are only allowed to participate in the construction of the regression function. In this way, the proposed $\epsilon$-SVQR model brings the sparsity back in the SVQR model.
                        Unlike Sparse SVQR, the proposed $\epsilon$-SVQR model can obtain major improvement in the prediction by tunning the width of the  $\epsilon$-tube. Extensive experiments with several artificial and real-world benchmark datasets show the efficacy of the proposed  $\epsilon$-SVQR model.


 The rest of this paper is organized as follows. Section-2 briefly describes the standard Support Vector Quantile Regression model\cite{quantile3} and Sparse Support Vector Quantile Regression model\cite{sparsequantile}. In Section-3, we  present the formulation of  proposed $\epsilon$-Support Vector Quantile Regression model. 
 Section-5 contains the numerical results obtained by  extensive experiments carried on  several artificial and UCI datasets. It empirically shows the advantages of the proposed SVQR model and other existing SVQR models. Section-6 concludes the main contribution of our work.

 \section{Support Vector Quantile Regression models}
 For the given training set $T$ and the quantile $\tau \in (0,1)$ , the  SVQR model estimates the conditional quantile function $f_{\tau}(x) = w^T\phi(x)+ b $ in the feature space, where $\phi:\mathbb{R}^n \rightarrow \mathcal{H}$ is a mapping from the input space to a higher dimensional feature space $\mathcal{H}$.
 \subsection{Standard Support Vector Quantile Regression model}
     The standard SVQR model minimizes
 \begin{eqnarray}
 \min_{w,b}~ \frac{1}{2}||w||^2 + C.\sum_{i=1}^{l}P_\tau({y_i-(w^Tx_i+b)}),
 \end{eqnarray}
 which can be equivalently converted to the following Quadratic Programming Problem (QPP)
\begin{eqnarray}
 \min_{(w,b,\xi,\xi^*)}~~ \frac{1}{2}||w||^2 + C.\sum_{i=1}^{l}(\tau\xi_i+ (1-\tau)\xi_i^{*}) \nonumber \\
  & \hspace{-105mm}\mbox{subject to,}\nonumber\\
 & \hspace{-70mm}y_i- (w^T\phi(x_i)+b) \leq   \xi_i,  \nonumber\\
 & \hspace{-70mm}(w^T\phi(x_i)+b)-y_i \leq  \xi_i^{*} , \nonumber\\
  & \hspace{-60mm}\xi_i \geq 0,~~\xi_i^{*} \geq 0, ~~~ i =1,2,...l.
  \label{SVQR_primal}
\end{eqnarray}
Here $C \geq 0$ is a user defined parameter which is used to find a good trade-off between empirical risk and flatness of the regressor.
  To solve the primal problem (\ref{SVQR_primal}) efficiently, we derive its corresponding Wolfe dual problem and obtained the following QPP
  \begin{eqnarray}
  \min_{(\alpha, \alpha^*)} ~\frac{1}{2} \sum_{i=1}^{l}\sum_{j=1}^{l}(\alpha_i- \alpha_j^*)K(x_i,x_j)(\alpha_j-\alpha_i^*) - \sum_{i=1}^{l}y_i(\alpha_i-\alpha_i^{*}) \nonumber \\
   & \hspace*{-180mm}\mbox{subject to,} \nonumber \\
& \hspace*{-140mm}  0 \leq \alpha_i \leq \tau C,  \nonumber \\
& \hspace*{-105mm} 0 \leq \alpha_i^{*} \leq (1-\tau) C, ~~~~~ i=~1,2,...l. 
\label{SVQR_dual}
        \end{eqnarray}
  
  After obtaining the optimal values of $\alpha_i$ and $\alpha_i^{*}$ from the dual problem (\ref{SVQR_dual}),  the quantile regression function $f_\tau(x) $, for any test data point $x \in \mathbb{R}^n$, is estimated as  
  
   \begin{equation}
   f_\tau(x) = \sum_{i=1}^{l}(\alpha_i-\alpha_i^{*})K(x,x_i) + b.
   \end{equation}
The value of bias $b$ can be computed by using the KKT conditions for the primal problem (\ref{SVQR_primal}) as is  done in the traditional $\epsilon$-SVR model.

\subsection{Sparse Support Vector Regression Model}
The Sparse SVQR model \cite{sparsequantile} uses the e-insensitive loss function $h_{\tau}(x)$ to measure the empirical risk along with the regularization term for the estimation of the quantile function. It seeks to find the solution of the optimization problem 
 \begin{eqnarray}
 \min_{w,b} \frac{1}{2}||w||^2 + C.\sum_{i=1}^{l}h_\tau({y_i-(w^Tx_i+b)}), \label{sparsesvqrprimal}
 \end{eqnarray}
which has been converted to the following constrained optimization problem  by Soek et al. in \cite{sparsequantile})
\begin{eqnarray}
 \min_{(w,b,\xi,\xi^*)} \frac{1}{2}||w||^2 + C.\sum_{i=1}^{l}(\tau\xi_i+ (1-\tau)\xi_i^{*}) \nonumber \\
  & \hspace{-110mm}\mbox{subject to,}\nonumber\\
 & \hspace{-70mm}y_i- (w^T\phi(x_i)+b) \leq \xi_i + 1-{\tau}^2 \epsilon,  \nonumber\\
 & \hspace{-70mm}(w^T\phi(x_i)+b)-y_i \leq  \xi_i^{*}+ \tau-\tau\epsilon , \nonumber\\
  & \hspace{-60mm}\xi_i \geq 0,~~\xi_i^{*} \geq 0, ~~~ i =1,2,...l.
  \label{Sparse_SVQR_primal}
\end{eqnarray}
Here $\epsilon \geq 0 $ and $C \geq 0$ are user defined parameters.  Like standard SVQR model, the Wolfe dual problem corresponding to the primal problem (\ref{Sparse_SVQR_primal}) can also be solved to obtain the desired quantile estimate.
  \section{Proposed $\epsilon$-Support Vector Quantile Regression model}
  
  In this section, we present our  proposed $\epsilon$-Support Vector Quantile Regression model ($\epsilon$-SVQR) model which uses the proposed asymmetric $\epsilon$-pinball loss function for measuring the empirical risk. The proposed $\epsilon$-SVQR model minimizes
   \begin{eqnarray}
 \min_{(w,b)} ~\frac{1}{2}||w||^2 + C. \sum_{i=1}^{l}L_{\tau}^{\epsilon}(y_i,x_i,w,b)  \nonumber \\
    & \hspace{-60mm}   =  \min_{(w,b)}~ \frac{1}{2}||w||^2 + C. \sum_{i=1}^{l}max(-(1-\tau)(y_i-(w^Tx_i+b)+\tau\epsilon),  \nonumber \\ & \hspace {-05mm}0,
    \tau(y_i-(w^Tx_i+b)-(1-\tau)\epsilon)) 
   \end{eqnarray}
   which can be  further rewritten as
     \begin{eqnarray}
    \min_{(w,b)} ~\frac{1}{2}||w||^2 + C\sum_{i=1}^{l}max(~-(1-\tau)(y_i-(w^Tx_i+b)+\tau\epsilon),~0~)\nonumber \\ &  \hspace{-110mm} ~+ ~ C\sum_{i=1}^{l}max(~0,~\tau(y_i-(w^Tx_i+b)-(1-\tau)\epsilon)~) \label{esvqr1}
    \end{eqnarray}
  After introducing the variables $\xi_i = max(~-(1-\tau)(y_i-(w^Tx_i+b)+\tau\epsilon),~0~)$ and $\xi^{*}_i = max(~0,~\tau(y_i-(w^Tx_i+b)-(1-\tau)\epsilon)~)$,  for $i~=1,2,...l$, the problem (\ref{esvqr1}) can be converted to  the following QPP
  \begin{eqnarray}
  \min_{(w,b,\xi,\xi^*)}~~ \frac{1}{2}||w||^2 + C.\sum_{i=1}^{l}(\xi_i+ \xi_i^{*}) \nonumber \\
    & \hspace{-95mm}\mbox{subject to,}\nonumber\\
   & \hspace{-50mm} \xi \geq  -(1-\tau)(y_i-(w^Tx_i+b)+\tau\epsilon) ,  \nonumber\\
   & \hspace{-50mm} \xi_i^{*} \geq \tau(y_i-(w^Tx_i+b)-(1-\tau)\epsilon) , \nonumber\\
    & \hspace{-60mm}\xi_i \geq 0,~~\xi_i^{*} \geq 0, ~~~ i =1,2,...l,
    \label{esvqr2}
  \end{eqnarray}
   which can be written in the standard form as follows
   \begin{eqnarray}
    \min_{(w,b,\xi,\xi^*)} ~~\frac{1}{2}||w||^2 + C.\sum_{i=1}^{l}(\xi_i+ \xi_i^{*}) \nonumber \\
     & \hspace{-95mm}\mbox{subject to,}\nonumber\\
    & \hspace{-50mm}y_i- (w^T\phi(x_i)+b) \leq (1-\tau)\epsilon +  \frac{\xi_i}{\tau},  \nonumber\\
    & \hspace{-57mm}(w^T\phi(x_i)+b)-y_i \leq   \tau\epsilon + \frac{\xi_i^{*}}{1-\tau} , \nonumber\\
     & \hspace{-60mm}\xi_i \geq 0,~~\xi_i^{*} \geq 0, ~~~ i =1,2,...l.
     \label{esvqr3}
   \end{eqnarray}
   After considering the replacement $\xi_i:= \frac{\xi_i}{\tau}$ and $\xi_i^{*}:=\frac{\xi_i^{*}}{1-\tau}$ in the above problem (\ref{esvqr3}), the primal problem of the proposed $\epsilon$-SVQR model is obtained as
   
   \begin{eqnarray}
       \min_{(w,b,\xi,\xi^*)} \frac{1}{2}||w||^2 + C.\sum_{i=1}^{l}(\tau\xi_i+ (1-\tau)\xi_i^{*}) \nonumber \\
        & \hspace{-110mm}\mbox{subject to,}\nonumber\\
       & \hspace{-50mm}y_i- (w^T\phi(x_i)+b) \leq (1-\tau)\epsilon +  \xi_i,  \nonumber\\
       & \hspace{-60mm}(w^T\phi(x_i)+b)-y_i \leq   \tau\epsilon + \xi_i^{*} , \nonumber\\
        & \hspace{-60mm}\xi_i \geq 0,~~\xi_i^{*} \geq 0, ~~~ i =1,2,...l.
        \label{esvqr_primal}
      \end{eqnarray}
      Here $\epsilon \geq 0$ is the user defined parameter.
   It is notable that with $\epsilon=0$ , the proposed $\epsilon$-SVQR model reduces to the SVQR model (Takeuschi et al., \cite{quantile3}).    
For solving the primal problem (\ref{esvqr_primal}) efficiently, we need to derive its  Wolfe dual problem. The Lagrangian function for the primal problem (\ref{esvqr_primal}) is obtained as
\begin{eqnarray}
 & \hspace{-160mm} L(w, b, \xi_i ,\xi_i^{*}, \alpha_i,\beta_i,\gamma_i,\lambda_i) = ~~~~ \frac{1}{2}||w||^2 + C.\sum_{i=1}^{l}(\tau\xi_i+ (1-\tau)\xi_i^{*})  \nonumber \\
+\sum_{i=1}^{l}\alpha_i(y_i- (w^T\phi(x_i)+b)- (1-\tau)\epsilon -  \xi_i)+ \sum_{i=1}^{l}\beta_i((w^T\phi(x_i)+b)-y_i - \tau\epsilon - \xi_i^{*})\nonumber \\   &\hspace{-220mm}-\sum_{i=1}^{l}\gamma_i\xi_i-
\sum_{i=1}^{l}\lambda _i\xi_i^{*}
\end{eqnarray}
  We can now note the KKT conditions for (\ref{esvqr_primal}) as follows
 \begin{eqnarray}
  & \hspace{-90mm}\frac{\partial L}{\partial w} = w+ \sum_{i=1}^{l}(\beta_i-\alpha_i)\phi(x_i)=0 \implies w = \sum_{i=1}^{l}(\alpha_i-\beta_i)\phi(x_i)  \label{kkt1}\\
  & \hspace{-152mm}\frac{\partial L}{\partial b}= \sum_{i=1}^{l}(\beta_i-\alpha_i) = 0. \label{kkt2}\\
  & \hspace{-134mm} \frac{\partial L}{\partial \xi_i}=  C\tau - \alpha_i -\gamma_i = 0,  ~~ i=1 ,2,...,l.\label{kkt3} \\
& \hspace{-125mm}\frac{\partial L}{\partial \xi_i^*}=  C(1-\tau) - \beta_i -\lambda_i = 0, ~~ i=1 ,2,...,l.\label{kkt4}\\
& \hspace{-105mm} \alpha_i(y_i- (w^T\phi(x_i)+b)- (1-\tau)\epsilon -  \xi_i) =0, ~~ i=1 ,2,...,l. \label{kkt5}\\
 \hspace{-8mm}\beta_i((w^T\phi(x_i)+b)-y_i - \tau\epsilon - \xi_i^{*}) = 0, ~~ i=1 ,2,...,l.~~~~~~~\label{kkt6}\\
& \hspace{-140mm}  \gamma_i\xi_i = 0,~~ \lambda _i\xi_i^{*} = 0~~ i=1 ,2,...,l.\label{kkt7}\\
 & \hspace{-115mm} y_i- (w^T\phi(x_i)+b) \leq (1-\tau)\epsilon +  \xi_i, i=1 ,2,...,l.\label{kkt8} \\
& \hspace{-118mm} (w^T\phi(x_i)+b)-y_i \leq   \tau\epsilon + \xi_i^{*}, ~~ i=1 ,2,...,l., \label{kkt9}\\
 &\hspace{ -140mm}\xi_i \geq 0,~~\xi_i^{*} \geq 0, ~~ i=1 ,2,...,l.\label{kkt10}
         \end{eqnarray}   
    Making the use the above KKT conditions, the Wolfe dual problem of the primal problem (\ref{esvqr_primal}) can be obtained as follows
     \begin{eqnarray}
    \min_{\alpha,\beta} \frac{1}{2}\sum_{i=1}^{l}\sum_{j=1}^{l}(\alpha_i- \beta_j)K(x_i,x_j)(\alpha_j-\beta_i) - \sum_{i=1}^{l}(\alpha_i-\beta_i)y_i + \sum_{i=1}^{l}((1-\tau)\epsilon\alpha_i+\tau\epsilon\beta_i) \nonumber\\
    & \hspace*{-240mm}\mbox{subject to,} \nonumber \\
    &\hspace{-190mm} \sum_{i=1}^{l}(\alpha_i-\beta_i)= 0, \nonumber \\
    & \hspace{-180mm}  0 \leq \alpha_i \leq C\tau,~~ i=~1,2,...l,       \nonumber \\
    & \hspace{-172mm} 0 \leq \beta_i \leq C(1-\tau), ~~i=~1,2,...l.  
    \label{esvqrdual}   
    \end{eqnarray}
   The KKT conditions (\ref{kkt1})- (\ref{kkt10}) will help us to discover the various characteristics of the proposed $\epsilon$-SVQR model. At first, we shall state following preposition.
   \newline

  \textbf{Preposition 1.}   For $\epsilon \geq 0,~ \alpha_i\beta_i$=0 holds $\forall$ $\textit{i=1,2,...l} $.\\ 
  
  Proof:- If possible, let us suppose there exists an index $i$ such that $\alpha_i\beta_i \neq 0$ holds. It implies that $\alpha_i\neq 0$ and $\beta_i \neq 0$. Therefore, from the KKT condition (\ref{kkt5}) and (\ref{kkt6}) we can obtain
  \begin{eqnarray}
  (y_i- (w^T\phi(x_i)+b)- (1-\tau)\epsilon -  \xi_i) =0  \label{11}\\
  \mbox{and}~~~~~~~~~~~~~~~~~~~~~~ ((w^T\phi(x_i)+b)-y_i - \tau\epsilon - \xi_i^{*}) =0. \label{22} 
  \end{eqnarray}
  Adding equation (\ref{11}) and (\ref{22}) gives
  $\xi_i^{*}+ \xi_i = -\epsilon$ which is  possible only when either $\xi_i \leq 0$ or $\xi_i^{*} \leq 0$. But, the KKT condition (\ref{kkt10}) requires $\xi_i \geq 0,~~\xi_i^{*} \geq 0, ~~for~ i=1 ,2,...,l.$ which contradicts our assumption. This proves the proposition.
  
  Further, let us locate the training points with the help of their obtained Lagrangian multipliers $\alpha_i$ and $\beta_i$ values. 
  
  For this, we consider the following three disjoint sets
  \begin{eqnarray}
  S_1=\{i: 0<\alpha_i \leq C\tau ~~or~~ 0<\beta_i \leq C (1-\tau)  \}, \nonumber \\
   & \hspace{-86 mm} S_2=\{i: \alpha_i = C\tau ~~or~~ \beta_i = C(1-\tau)  \}, \nonumber \\
   & \hspace{-96 mm} S_3=\{i: \alpha_i = 0 ~~and~~  \beta_i = 0 \}. \nonumber 
  \end{eqnarray}

 For all training data points $x_i \in S_1$ we will  have $\gamma_i > 0$ (or $\lambda_i >0$) from KKT condition (\ref{kkt3}) (or (\ref{kkt4})). It implies that $\xi_i=0$ (or $\xi_i^*=0$)  from KKT condition (\ref{kkt7}). But since, $\alpha_i \geq 0$ (or  $\beta_i \geq 0$) so  condition (\ref{11}) (or (\ref{22})) must satisfy which will consequently imply that
 \begin{eqnarray}
   (y_i- (w^T\phi(x_i)+b)- (1-\tau)\epsilon ) =0   \label{112}\\
   (\mbox{or}~~~~~~~ ((w^T\phi(x_i)+b)-y_i - \tau\epsilon ) =0) \label{122} 
   \end{eqnarray}
 will hold true. It means that all training data point $x_i \in S_1$ will be located on the boundary points  of the $\epsilon$-insensitive zone. 
        Further, the data point which satisfies $0<\alpha_i \leq C\tau$ will be lying on the upper boundary of the asymmetric $\epsilon$-insensitive zone.  The data point which satisfies $0<\beta_i \leq C(1-\tau)$ will be lying on the lower boundary of the asymmetric $\epsilon$-insensitive zone. 
 
  For all training data points $x_i \in S_2$, we will have $\gamma_i = 0$ (or $\lambda_i  = 0) $ from KKT condition (\ref{kkt3}) (or (\ref{kkt4})). It implies that $\xi_i \geq 0$ (or $\xi^{*} \geq 0$).  Using the KKT condition  (\ref{kkt5}) ( or (\ref{kkt6})) we can obtain
  \begin{eqnarray}
  y_i- (w^T\phi(x_i)+b) > (1-\tau)\epsilon    
  ~~(~\mbox{or}~~ (w^T\phi(x_i)+b)-y_i   > \tau\epsilon~~) \label{123} 
  \end{eqnarray}
  
  It means that these training data points are lying outside of the asymmetric $\epsilon$-insensitive zone. Further, the data point for which $ \alpha_i = C\tau $  will be lying above the asymmetric $\epsilon$-insensitive zone. The data point for which $ \alpha_i = C(1-\tau) $  will be lying below of the asymmetric $\epsilon$-insensitive zone. 
    
   For all training data point $x_i \in S_3$ will lie inside of the $\epsilon$-insensitive zone and will not contribute to errors. These data points are ignored and doesn't contribute in the construction of the regression function.
   
   Like the $\epsilon$-SVR model, the data points which are lying outside of the $\epsilon$-insensitive zone as well on the boundary of the $\epsilon$-insensitive zone only contribute to the estimated regressor. But, the proportion of their contributions is not equal in the proposed $\epsilon$-QSVR model. It depends upon the location of the data point as well as $\tau$ value. In the proposed $\epsilon$-SVQR model, the data point which lies above and below of the asymmetric $\epsilon$-insensitive zone contribute to the final quantile regressor in the ratio of $(1-\tau)$ and $\tau$.     For example, for $\tau < 0.5$ , the data point lying below of the  asymmetric $\epsilon$-insensitive zone are more important than data point lying above  of the   asymmetric $\epsilon$-insensitive zone in the construction of the regressor. It is only because of the fact that for $\tau < 0.5$,  few data points will be lying below of the asymmetric $\epsilon$-insensitive zone and more data points will be lying above of the asymmetric $\epsilon$-insensitive zone.    
   
   After obtaining the solution of the dual problem (\ref{esvqrdual}),  the quantile regression function $f_\tau(x) $, for any test data point $x \in \mathbb{R}^n$, is estimated as  
     
      \begin{equation}
      f_\tau(x) = \sum_{i=1}^{l}(\alpha_i-\beta_i)K(x,x_i) + b.
      \end{equation}
    For obtaining the optimal value of the bias term $b$, we can pick up the training data points in $S_1$ and can compute the value of b from the equation (\ref{112}) or (\ref{122}). In practice,  for every $\alpha_i > 0$, we compute the value of $b$ form equation (\ref{112}) and  for every $\beta_i > 0$,  we compute the value of $b$ form equation (\ref{122}) and use the average of these values as final value of $b$. Further, like $\epsilon$-SVR model discribed in (Gunn, \cite{GUNNSVM}), if the kernel contains a bias term then, the $\epsilon$-SVQR dual problem (\ref{esvqrdual}) can be solved without equality constraint and  the quantile regression function is simply estimated by
     \begin{equation}
    f_\tau(x) = \sum_{i=1}^{l}(\alpha_i-\beta_i)K(x,x_i) 
    \label{r12}
    \end{equation}

 \section{Experimental Results}
 In this section, we have performed extensive experiments to verify the efficacy of the proposed $\epsilon$-SVQR model. For this, we first describe our experimental setup. We have performed all experiments with MATLAB 16.0 environment (http://in.mathworks.com/) on Intel XEON processor with 16.0 GB of RAM. Since the proposed $\epsilon$-SVQR  model is  basically an improvement over the standard SVQR model, so we shall only consider existing SVQR models for experiments. The QPPs of proposed $\epsilon$-SVQR and Sparse SVQR has been solved by the quadprog function with interior-point convex algorithm available in the MATLAB 16.0 environment. It is also noteworthy that the SVQR model is a special case  of the proposed $\epsilon$-SVQR and Sparse SVQR model with a particular choice of $\epsilon=0$.
  For all of the experiments, we have used the RBF kernel function $exp(\frac{-||x-y||^2}{q})$, where $q$ is the kernel parameter and quantile regression function is estimated by  (\ref{r12}). 
                                        The proposed $\epsilon$-SVQR model and Sparse SVQR model involves three parameters namely RBF kernel parameter $q$,$C$ and $\epsilon$. These parameters have been tunned with exhaustive search method (Hsu and Lin, \cite{Exhaustivesearch}). The parameter $q$ and $C$ has been searched in the set $\{ 2^i: i=-15,-9,......9,15\} $. The parameter $\epsilon$ has been searched in the set $\{ 0,0.1,0.2,...2,2.5,3..,5\}$.

   \subsection{\textbf{Performance Criteria}}  \label{Perform_criteria}                                   
    For comparison of the efficacy of  SVQR  models, we have used some evaluation criteria which is also mentioned in (Xu Q et al., \cite{Weighted_QSVR}).
      Given the training set $T= \{ (x_i,y_i): x_i \in \mathbb{R}^n, y_i \in \mathbb{R},~ i=1,2...,l~ \}$  and true  $\tau$-th conditional quantile function $Q_{\tau}(y/x)$, we list the evaluation criteria as follows.  
     \begin{enumerate}
     \item[(i)] $ RMSE$: It is Root Mean Square of Error.\\ ~~It is given by $\sqrt{ \frac{1}{l}\sum_{i=1}^{l}( Q_{\tau}(y_i/x_i)- f_\tau(x_i))^2}$.
          \item[(ii)] $ MAE$: It is  Mean of the Absolute Error. \\
          ~~~It is given by ${ \frac{1}{l}\sum_{i=1}^{l}|( Q_{\tau}(y_i/x_i)- f_\tau(x_i))|}$.
          \item[(iii)] $ TheilU$: It is a measure used for the quantile regression estimate. It is given by  $\sqrt{ \frac{\frac{1}{l}\sum_{i=2}^{l}( Q_{\tau}(y_i/x_i)- f_\tau(x_i))^2 / Q_{\tau}(y_i/x_i) 
           }   { (\frac{1}{l}\sum_{i=2}^{l}( Q_{\tau}(y_{i-1}/x_i)- f_\tau(x_i)))^2 / Q_{\tau}(y_i/x_i)}}$. 
           If its value is less than 1 then the used quatile regression is better than guessing.
           \item[(iv)]  Error $E_\tau$: It is the measure which is used when the true quantile function is unknown.
              It is given by $E_\tau ~=~ |p_\tau -\tau|$, where  $p_{\tau} = P(y_i \leq f_{\tau}(x_i))$ is the coverage probablity. For the real world UCI datasets  experiments, we would be using this measure. We shall compute the coverage probability $p_{\tau}$ by obtaining the estimated $\tau$ value in 100 random trails.
           \item[(v)]  Sparsity(u) = $\frac{\#(u=0)}{\#(u)}$, where $\#(r)$ determines the number of the component of the vector $r$ .
     \end{enumerate}

\subsection{Artifical Datasets}
\begin{figure}
 	\centering
 	\subfloat[] {\includegraphics[width=2.5in,height=1.5in]{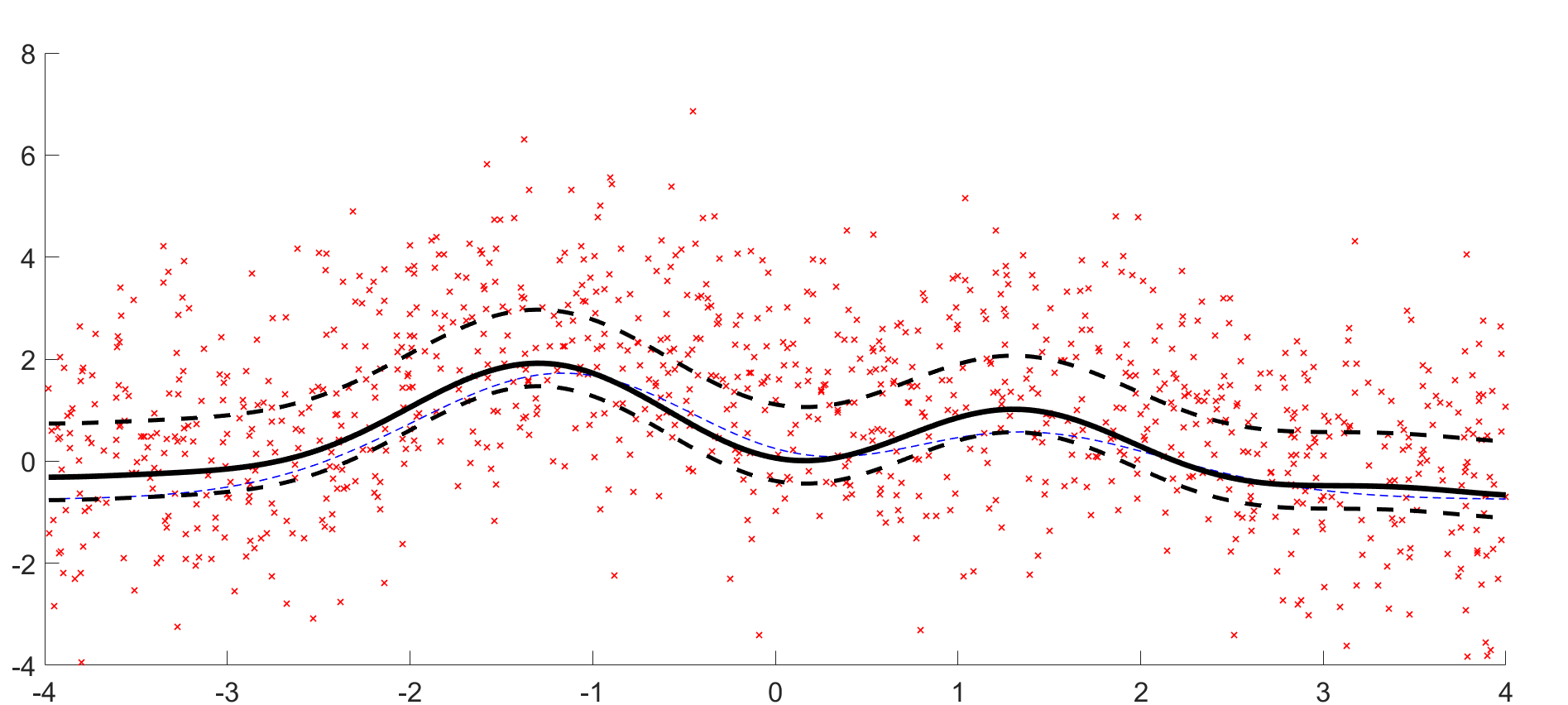}}
 	\subfloat[] {\includegraphics[width=2.5in,height=1.5in]{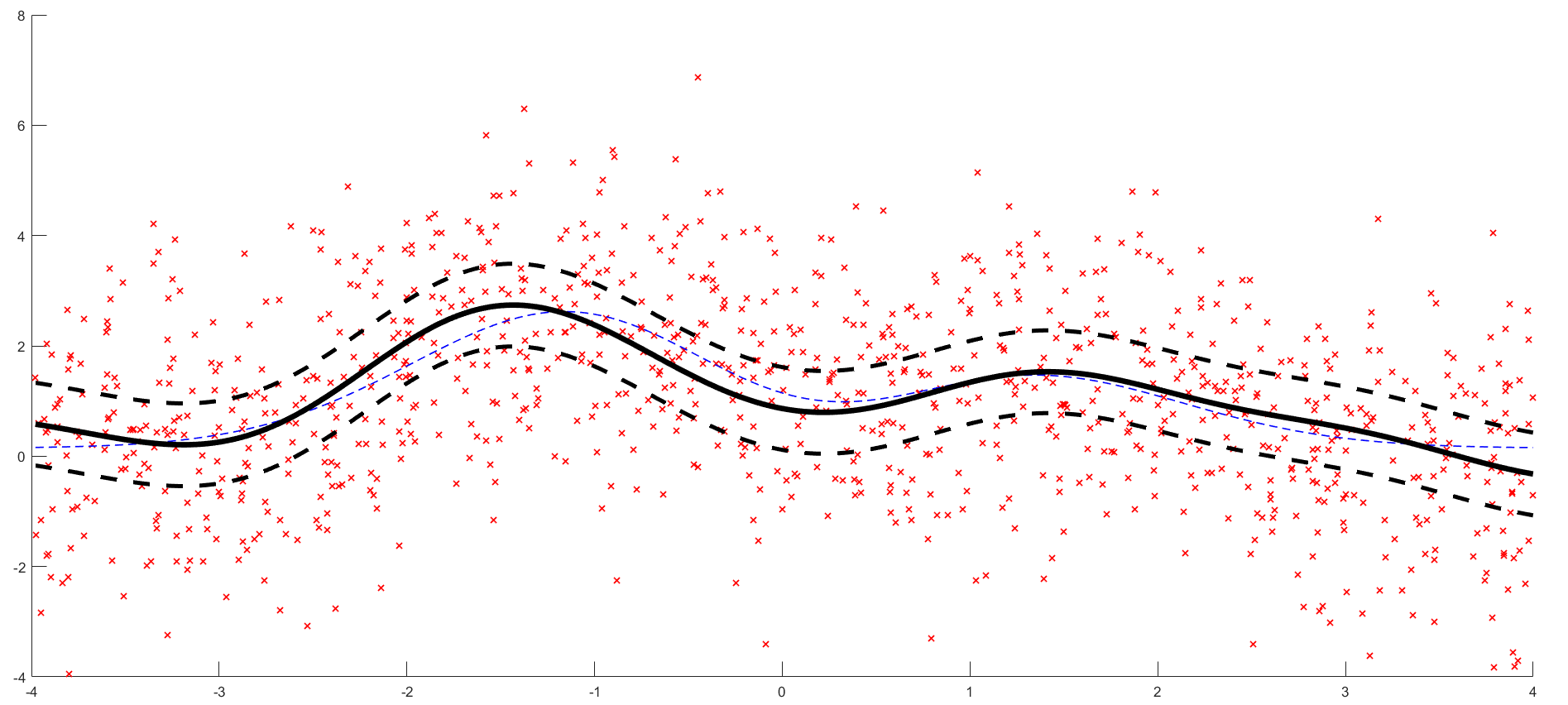}}\\
 	\subfloat[] {\includegraphics[width=2.5in,height=1.5in]{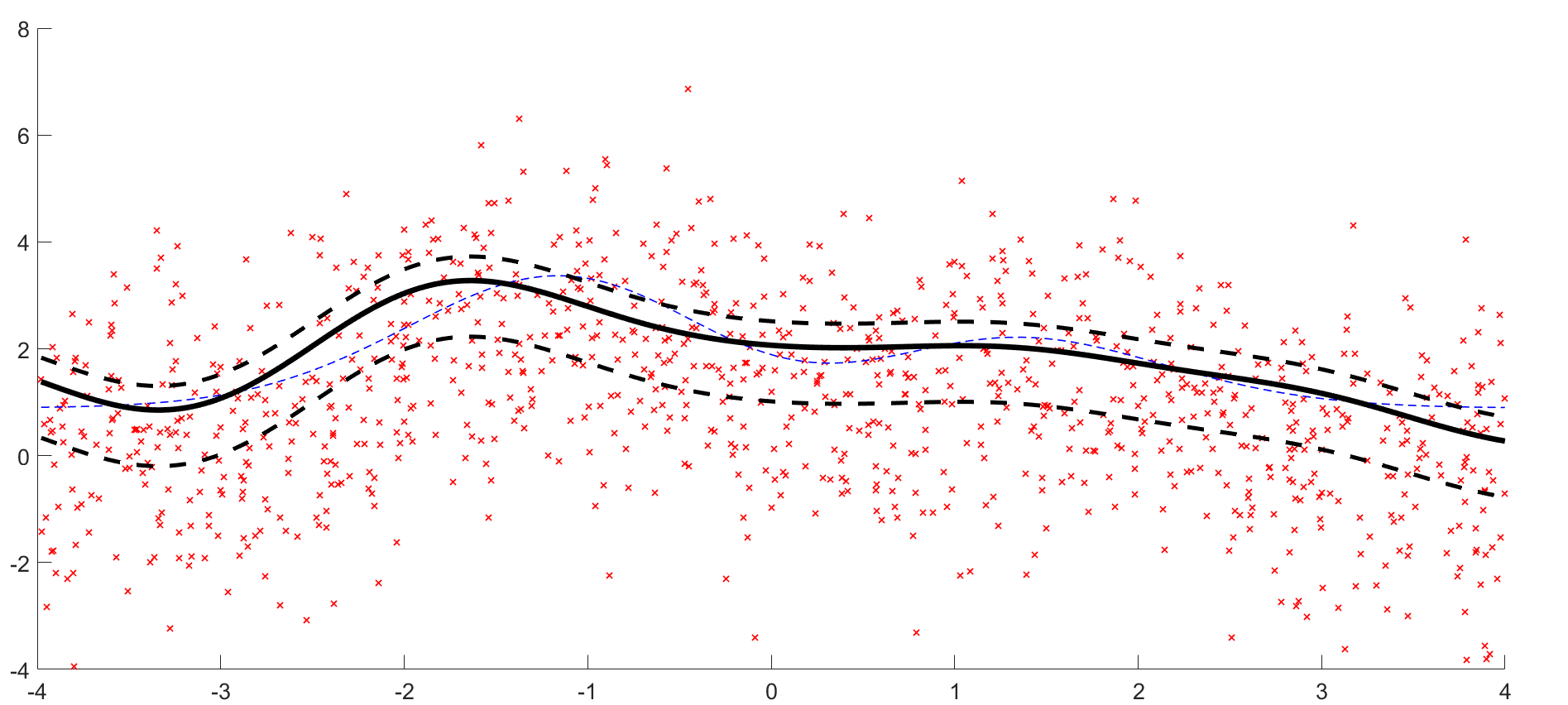}}
 	\subfloat[] {\includegraphics[width=2.5in,height=1.5in]{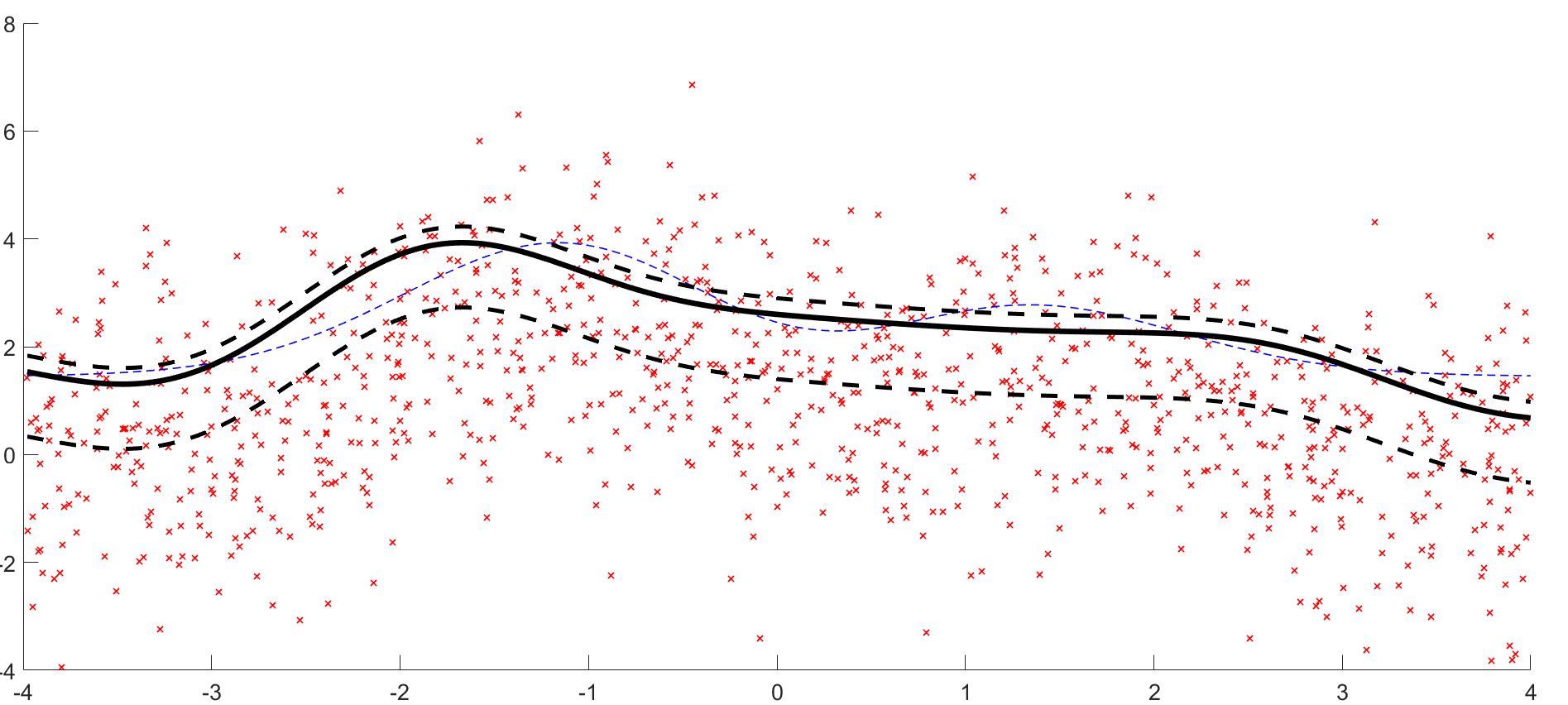}}
 	\caption{Performance of the proposed $\epsilon$-QSVR model with $\epsilon=1.5$ for (a) $\tau=0.3$ (b) $\tau=0.5$  (c) $\tau=0.7$ and  (d) $\tau=0.8$}
 	\label{plot2}
 \end{figure}

\begin{figure}
 	\centering
 	\subfloat[] {\includegraphics[width=0.6\linewidth]{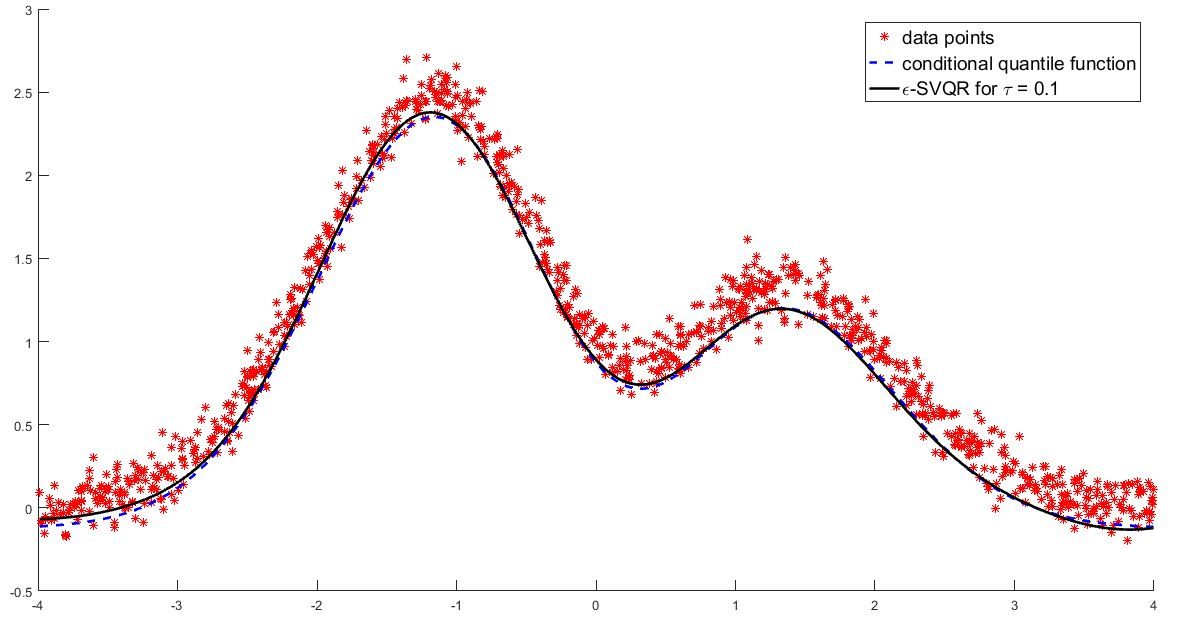}}
 	\subfloat[] {\includegraphics[width=0.6\linewidth]{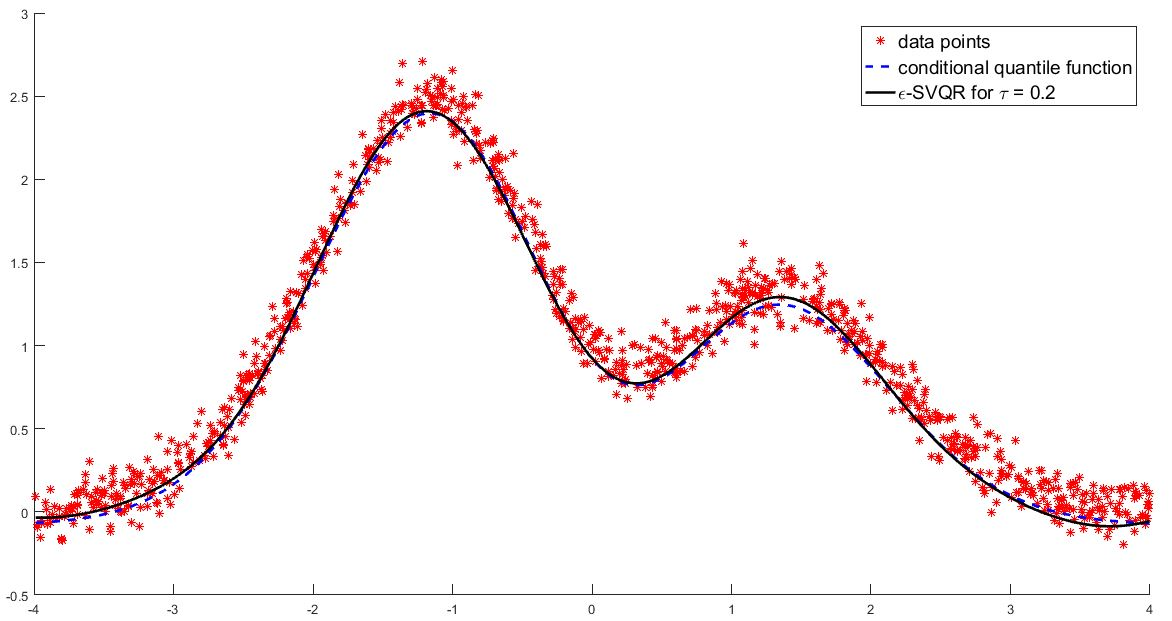}}\\
 	\subfloat[] {\includegraphics[width= 0.6\linewidth]{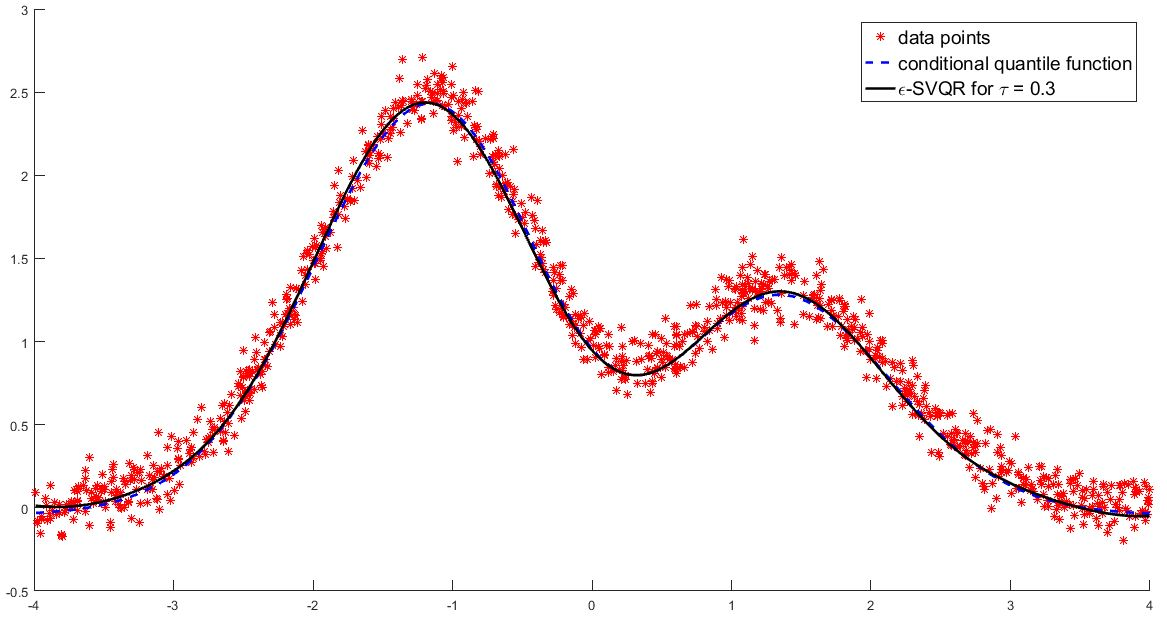}}
 	\subfloat[] {\includegraphics[width=0.6\linewidth]{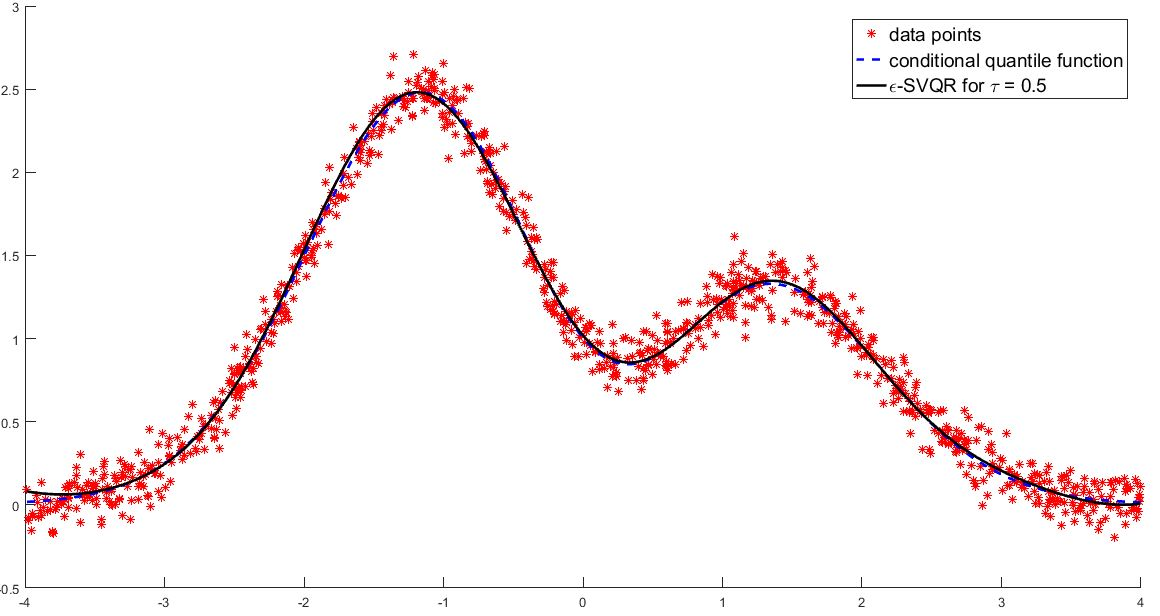}}\\
 	\subfloat[] {\includegraphics[width=0.6\linewidth]{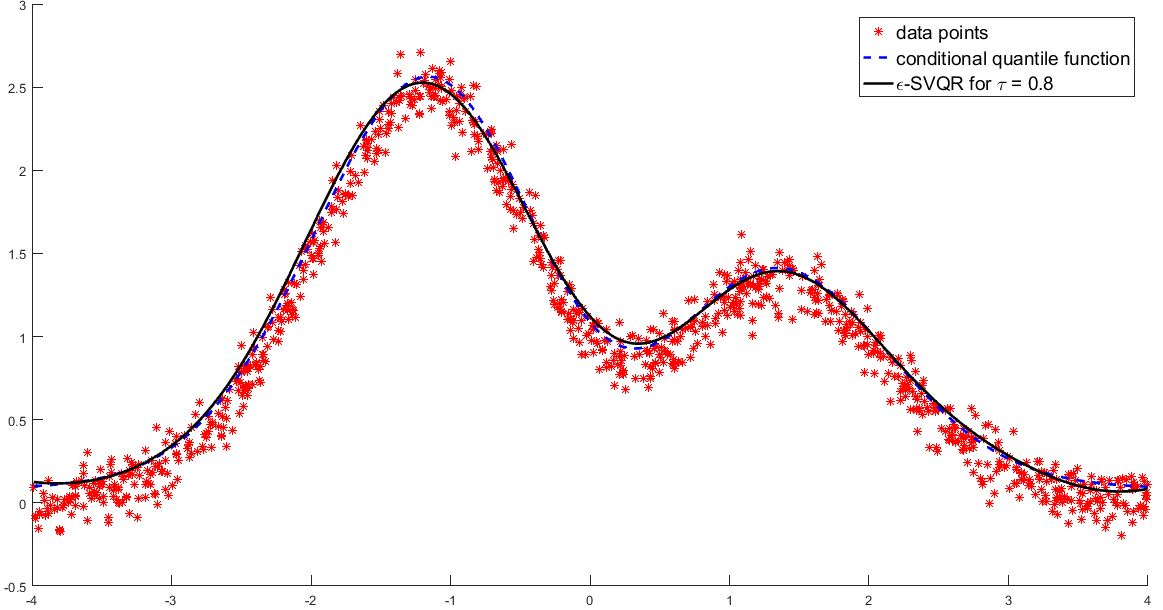}}
 	\subfloat[] {\includegraphics[width=0.6\linewidth]{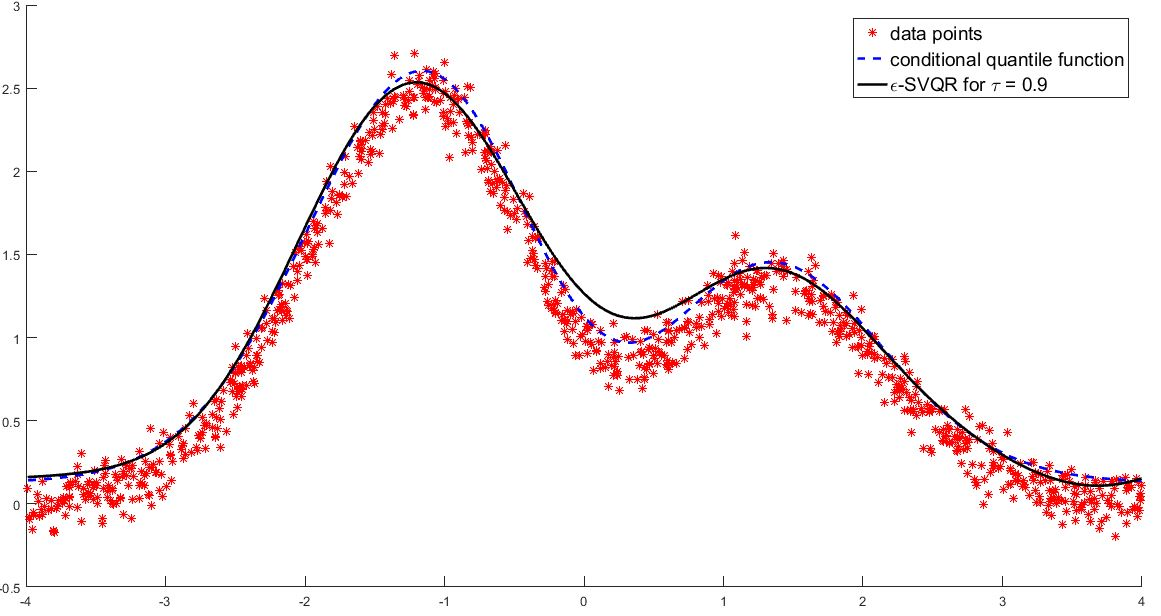}}
 	\caption{One run simulation of the proposed $\epsilon$-QSVR model with different $\tau$ values.}
 	\label{plot}
 \end{figure}

We need to observe the role of $\epsilon$-insensitive zone in SVQR models and prove the efficacy of the proposed $\epsilon$-SVQR model over Sparse SVQR model empirically. For this, we have considered artificial datasets with different nature of noises. We have generated the training set $T$ where $x_i$ is drawn from the univariate uniform distribution with  $[-4,4]$.  The response variable $y_i$ is obtained from polluting a nonlinear function  of $x_i$ with different natures of noises in artificial datasets as follows.

\begin{eqnarray}
& \hspace{5mm} \mbox{AD1:}~~y_i = (1-x_i+2x_i^{2})e^{-0.5x_{i}^2} + \xi_i, \nonumber
\mbox{~~~~~where $\xi_i$ is from N(0, $\sigma$).}  \nonumber\\
& \hspace{0mm} \mbox{AD2:}~~y_i = (1-x_i+2x_i^{2})e^{-0.5x_{i}^2} + \xi_i,
\mbox{~~~~~where $\xi_i$ is from} ~\chi^2(3) \nonumber
\end{eqnarray}
 The artificial datasets AD1 and AD2 contain 200 training points. The true quantile function $Q_\tau(y_i/x_i)$ in these artificial datasets can be obtained as
 \begin{eqnarray}
y_i = (1-x_i+2x_i^{2})e^{-0.5x_{i}^2} + F_{\tau}^{-1}(\xi_i), \nonumber
 \end{eqnarray}  
 where $F_{\tau}^{-1}(\xi_i)$ is the  $\tau$th quantile of random error $\xi_i$. We have evaluated the SVQR models in 100 independent trails by generating 1000 testing points in each trails.  The one run simulation of the artificial dataset AD1 with $\sigma =0.2$ has been plotted in the Figure \ref{plot} along with the true quantile estimates and predicted quantile estimates by proposed $\epsilon$-SVQR model for several $\tau$ values. Figure (\ref{plot2}) shows the performance of the proposed $\epsilon$-SVQR model with $\epsilon = 1.5$ on  artificial AD1 dataset with  $\sigma= 1.5$ for different $\tau$ values. It also shows that how proposed asymmetric $\epsilon$-insensitive pinball loss function can obtain the asymmetric $\epsilon$-insensitive zone around the data with different $\tau$ values. The width of the  $\epsilon$-insensitive zone remain fixed for different $\tau$ values but its division varies with $\tau$ values. It enables the proposed $\epsilon$-SVQR model to obtain better estimate irrespective of $\tau$ values.
 
    To show the efficacy of proposed $\epsilon$-SVQR model over existing Sparse SVQR model, we have tested both of them on artificial dataset AD1 with different noise variances $\sigma$. We have listed the RMSE values obtained by $\epsilon$-SVQR model and Sparse SVQR model with different $\sigma$ and $\epsilon$ values for $\tau$= 0.1 in Table \ref{table1} and Table \ref{table2} respectively. We can infer following observations from these Tables.
    
\begin{figure}
	\centering
	\includegraphics[width=1.0\linewidth, height= 0.3\textheight]{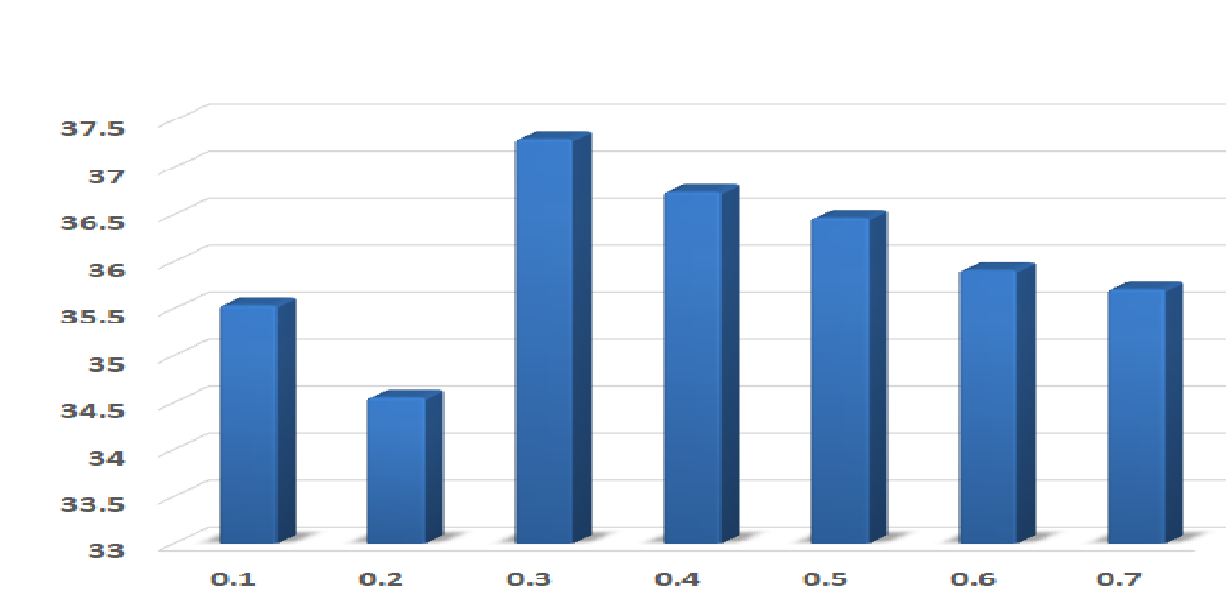}
	\caption{Percentage of improvement in RMSE  obtained by proposed $\epsilon$-SVQR over SVQR model after tunning  its parameter $\epsilon$ for different variance of noise $\sigma$ for $\tau=0.1$ }
	\label{plot112}
\end{figure}

\begin{figure}
\centering
\includegraphics[width=0.8\linewidth]{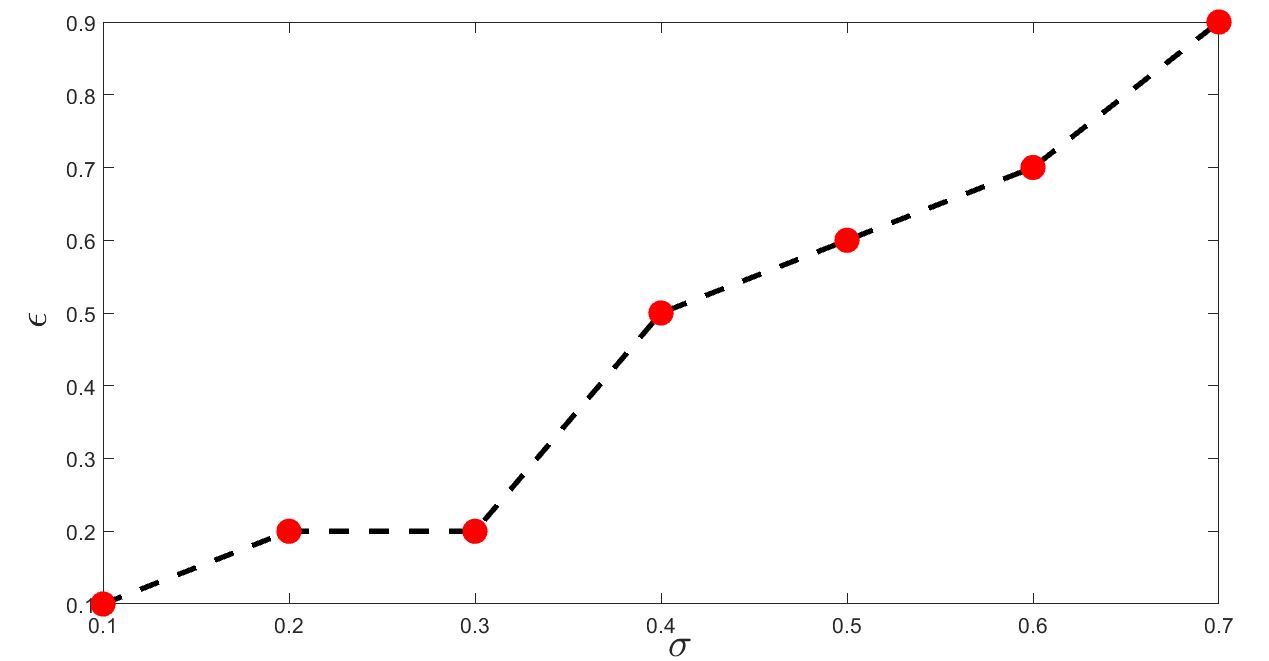}
\caption{ Plot of optimal $\epsilon$ values corresponding to minimum RMSE obtained by proposed $\epsilon$-SVQR model with $\sigma$ values for $\tau =0.1$. }
\label{sigmavseps11}
\end{figure}

    \begin{enumerate}
    \item[(i)] The  numerical values listed in the first column of Table \ref{table1} and Table \ref{table2} are same i,e. the Sparse SVQR model and $\epsilon$-SVQR model obtain same RMSE values with $\epsilon~=0$. It is because of fact that the both Sparse SVQR model and $\epsilon$-SVQR models reduce to the standard SVQR model with the value of $\epsilon$=0. Further, it can also be observed that the proposed $\epsilon$-SVQR model obtains better RMSE values on non zeros values of $\epsilon$. Also, Sparse SVQR obtains better RMSE values on non zeros values of $\epsilon$ in several cases. It confirms that the concept of the $\epsilon$ -insensitive zone is quite relevant in the SVQR models.
    
    \item[(ii)] It can be observed from RMSE values listed in the Table \ref{table1} that the proposed $\epsilon$-SVQR model can obtain major improvement over SVQR model by tuning its parameter value $\epsilon$. Figure (\ref{plot112}) shows the percentage of improvement in RMSE obtained by proposed $\epsilon$-SVQR model over standard SVQR model.  Also, the optimal choice of $\epsilon$ increases along with values of  noise variance $\sigma$ present in the responses of training dataset in proposed in $\epsilon$-SVQR model. This fact is also well depicted in the Figure \ref{sigmavseps11}. 
    
    \item[(iii)]The Sparse SVQR model struggles to obtain a good RMSE values with the given range of $\epsilon$ values. It is because of the fact that the width of the $\epsilon$-insensitive zone in the Sparse SVQR model does depend on the $\tau$ value. Further, it can also be observed that, the Sparse SVQR model obtains its optimal RMSE values at $\epsilon$=0 with the noise variance $\sigma$ =0,0.1,0.2,0.3 ,0.4. It means that the Sparse SVQR may fail to utilize the concept of the $\epsilon$-insensitive zone in certain cases.
    \end{enumerate}
      
      We have listed the optimal RMSE values obtained by the Sparse SVQR model and proposed $\epsilon$-SVQR model with different values of the noise variance $\sigma$ for the $\tau$ = 0.1 and $\tau$ = 0.9 in Table \ref{table3} and   Table \ref{table4} respectively  . It can observed that in most of cases, the proposed  $\epsilon$-SVQR obtains better RMSE values than existing Sparse SVQR model. 
 
\begin{table}[h]
{\footnotesize \begin{tabular}{|c|c|c|c|c|c|c|c|c|c|c|c|}
	\hline
$\sigma/\epsilon$ & 0.0   & 0.1   & 0.2   & 0.3   & 0.4   & 0.5   & 0.6   & 0.7   & 0.8   & 0.9 & 1  \\ \hline
0.1        & 0.043 & 0.027 & 0.035 & 0.058 & 0.077 & 0.116 & 0.165 & 0.214 & 0.264 & 0.311 & 0.365 \\ \hline
0.2        & 0.083 & 0.055 & 0.054 & 0.059 & 0.071 & 0.082 & 0.113 & 0.134 & 0.148 & 0.187 & 0.230 \\ \hline
0.3        & 0.124 & 0.100 & 0.077 & 0.080 & 0.083 & 0.094 & 0.107 & 0.111 & 0.134 & 0.167 & 0.184 \\ \hline
0.4        & 0.165 & 0.134 & 0.113 & 0.103 & 0.106 & 0.104 & 0.122 & 0.129 & 0.144 & 0.146 & 0.158 \\ \hline
0.5        & 0.204 & 0.174 & 0.153 & 0.128 & 0.129 & 0.131 & 0.129 & 0.144 & 0.156 & 0.155 & 0.177 \\ \hline
0.6        & 0.243 & 0.214 & 0.202 & 0.167 & 0.155 & 0.155 & 0.157 & 0.154 & 0.165 & 0.178 & 0.194 \\ \hline
0.7        & 0.282 & 0.256 & 0.232 & 0.210 & 0.185 & 0.182 & 0.184 & 0.184 & 0.182 & 0.181 & 0.200 \\ \hline
\end{tabular}}
\caption{RMSE obtained by the proposed $\epsilon$-SVQR model with different $\sigma$ and $\epsilon$ values for $\tau$=0.1. }
\label{table1}
\end{table}
\begin{figure}
	\centering
	\includegraphics[width=1.0\linewidth]{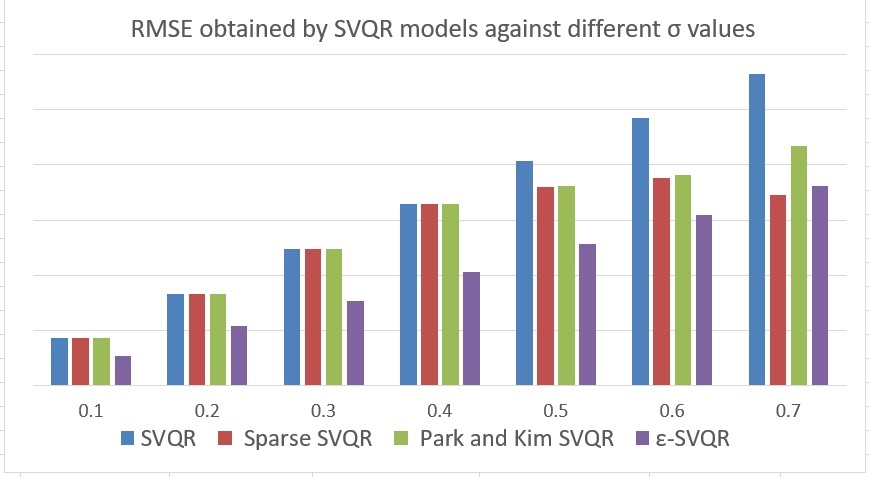}
	\caption{Comparisons of minimum RMSE obtained by proposed $\epsilon$-SVQR with other SVQR models after tunning  its parameter $\epsilon$ for different variance of noise $\sigma$ for $\tau=0.1$ } 
	\label{comp_svqrs_tau01}
\end{figure}

\begin{table}[h]
{\footnotesize \begin{tabular}{|c|c|c|c|c|c|c|c|c|c|c|c|}
\hline
$\sigma \setminus\epsilon$ & 0    & 0.1  & 0.2  & 0.3  & 0.4  & 0.5  & 0.6  & 0.7  & 0.8  & 0.9  & 1    \\ \hline
0.1        & 0.043 & 0.369 & 0.832 & 1.108 & 1.108 & 1.107 & 1.106 & 1.106 & 1.105 & 1.105 & 1.105 \\ \hline
0.2        & 0.083 & 0.242 & 0.701 & 1.029 & 1.028 & 1.027 & 1.026 & 1.025 & 1.024 & 1.023 & 1.022 \\ \hline
0.3        & 0.124 & 0.208 & 0.588 & 0.955 & 0.953 & 0.952 & 0.951 & 0.950 & 0.949 & 0.948 & 0.947 \\ \hline
0.4        & 0.165 & 0.176 & 0.466 & 0.847 & 0.893 & 0.892 & 0.890 & 0.888 & 0.887 & 0.886 & 0.884 \\ \hline
0.5        & 0.204 & 0.180 & 0.406 & 0.708 & 0.835 & 0.833 & 0.832 & 0.830 & 0.828 & 0.827 & 0.825 \\ \hline
0.6        & 0.243 & 0.189 & 0.408 & 0.609 & 0.787 & 0.784 & 0.782 & 0.780 & 0.777 & 0.775 & 0.773 \\ \hline
0.7        & 0.282 & 0.173 & 0.345 & 0.579 & 0.750 & 0.747 & 0.744 & 0.742 & 0.739 & 0.737 & 0.734 \\ \hline
\end{tabular}}
\caption{RMSE obtained by the Sparse SVQR model with different $\sigma$ and $\epsilon$ values for $\tau$=0.1. }
\label{table2}
\end{table}
   \begin{table}[h]
   	{\footnotesize \begin{tabular}{|c|c|c|c|c|c|c|c|c|c|c|c|}
   			\hline
   			$\sigma \setminus\epsilon$ & 0    & 0.1  & 0.2  & 0.3  & 0.4  & 0.5  & 0.6  & 0.7  & 0.8  & 0.9  & 1    \\ \hline
   			0.1        & 0.043 & 0.425 & 0.911 & 1.103 & 1.103 & 1.103 & 1.103 & 1.103 & 1.103 & 1.103 & 1.103 \\ \hline
   			0.2        & 0.083 & 0.289 & 0.774 & 1.014 & 1.013 & 1.013 & 1.013 & 1.013 & 1.013 & 1.013 & 1.013 \\ \hline
   			0.3        & 0.124 & 0.203 & 0.647 & 0.934 & 0.932 & 0.932 & 0.932 & 0.932 & 0.932 & 0.932 & 0.932 \\ \hline
   			0.4        & 0.165 & 0.194 & 0.519 & 0.862 & 0.857 & 0.855 & 0.857 & 0.860 & 0.862 & 0.862 & 0.862 \\ \hline
   			0.5        & 0.204 & 0.180 & 0.416 & 0.792 & 0.798 & 0.792 & 0.790 & 0.791 & 0.795 & 0.803 & 0.805 \\ \hline
   			0.6        & 0.243 & 0.191 & 0.380 & 0.651 & 0.745 & 0.742 & 0.737 & 0.736 & 0.738 & 0.744 & 0.752 \\ \hline
   			0.7        & 0.282 & 0.217 & 0.399 & 0.552 & 0.696 & 0.698 & 0.698 & 0.694 & 0.693 & 0.697 & 0.704 \\ \hline
   	\end{tabular}}
   	\caption{RMSE obtained by the Park and Kim SVQR model with different $\sigma$ and $\epsilon$ values for $\tau$=0.1. }
   	\label{table3}
   \end{table}


\begin{table}[h]
	{\footnotesize \begin{tabular}{|c|c|c|c|c|c|c|c|c|c|c|c|}
			\hline
			$\sigma/\epsilon$ & 0.0   & 0.1   & 0.2   & 0.3   & 0.4   & 0.5   & 0.6   & 0.7   & 0.8   & 0.9 & 1  \\ \hline
			0.1 & 0.059 & 0.055 & 0.046 & 0.064 & 0.077 & 0.104 & 0.114 & 0.142 & 0.176 & 0.209 & 0.207 \\ \hline
			0.2 & 0.112 & 0.119 & 0.109 & 0.090 & 0.091 & 0.105 & 0.118 & 0.134 & 0.142 & 0.162 & 0.160 \\ \hline
			0.3 & 0.166 & 0.169 & 0.171 & 0.159 & 0.141 & 0.136 & 0.137 & 0.151 & 0.163 & 0.175 & 0.189 \\ \hline
			0.4 & 0.213 & 0.223 & 0.221 & 0.226 & 0.213 & 0.193 & 0.182 & 0.183 & 0.183 & 0.196 & 0.205 \\ \hline
			0.5 & 0.263 & 0.279 & 0.259 & 0.263 & 0.259 & 0.257 & 0.250 & 0.235 & 0.232 & 0.227 & 0.234 \\ \hline
			0.6 & 0.317 & 0.330 & 0.307 & 0.290 & 0.293 & 0.299 & 0.299 & 0.299 & 0.289 & 0.276 & 0.279 \\ \hline
			0.7 & 0.372 & 0.372 & 0.347 & 0.337 & 0.339 & 0.345 & 0.344 & 0.339 & 0.346 & 0.344 & 0.335 \\ \hline
	\end{tabular}}
	\caption{RMSE obtained by the proposed $\epsilon$-SVQR model with different $\sigma$ and $\epsilon$ values for $\tau$=0.9. }
	\label{table11}
\end{table}
   
\begin{table}[h]
	{\footnotesize \begin{tabular}{|c|c|c|c|c|c|c|c|c|c|c|c|}
			\hline
			$\sigma \setminus\epsilon$ & 0    & 0.1  & 0.2  & 0.3  & 0.4  & 0.5  & 0.6  & 0.7  & 0.8  & 0.9  & 1    \\ \hline
		0.1    & 0.059  & 0.243  & 0.237  & 0.232  & 0.226  & 0.221  & 0.217  & 0.213  & 0.209  & 0.207  & 0.204 \\ \hline
		0.2    & 0.112  & 0.193  & 0.223  & 0.217  & 0.212  & 0.207  & 0.203  & 0.199  & 0.196  & 0.193  & 0.190 \\ \hline
		0.3    & 0.166  & 0.214  & 0.264  & 0.258  & 0.251  & 0.246  & 0.240  & 0.235  & 0.231  & 0.226  & 0.223 \\ \hline
		0.4    & 0.213  & 0.231  & 0.314  & 0.318  & 0.312  & 0.305  & 0.299  & 0.293  & 0.287  & 0.282  & 0.277 \\ \hline
		0.5    & 0.263  & 0.258  & 0.359  & 0.373  & 0.366  & 0.359  & 0.352  & 0.345  & 0.339  & 0.333  & 0.327 \\ \hline
		0.6    & 0.317  & 0.294  & 0.400  & 0.439  & 0.431  & 0.424  & 0.417  & 0.410  & 0.403  & 0.397  & 0.390 \\ \hline
		0.7    & 0.372  & 0.353  & 0.427  & 0.490  & 0.493  & 0.485  & 0.478  & 0.471  & 0.464  & 0.457  & 0.450 \\ \hline
	\end{tabular}}
	\caption{RMSE obtained by the Sparse SVQR model with different $\sigma$ and $\epsilon$ values for $\tau$=0.9. }
	\label{table22}
\end{table}   
   
  \begin{table}[h]
  	{\footnotesize \begin{tabular}{|c|c|c|c|c|c|c|c|c|c|c|c|}
  			\hline
  			$\sigma \setminus\epsilon$ & 0    & 0.1  & 0.2  & 0.3  & 0.4  & 0.5  & 0.6  & 0.7  & 0.8  & 0.9  & 1    \\ \hline
  	0.1 & 0.059 & 0.204 & 0.198 & 0.223 & 0.271 & 0.332 & 0.398 & 0.469 & 0.543 & 0.618 & 0.212 \\ \hline
  	0.2 & 0.112 & 0.190 & 0.198 & 0.237 & 0.294 & 0.361 & 0.433 & 0.509 & 0.586 & 0.665 & 0.171 \\ \hline
  	0.3 & 0.166 & 0.198 & 0.215 & 0.257 & 0.317 & 0.381 & 0.452 & 0.528 & 0.606 & 0.686 & 0.207 \\ \hline
  	0.4 & 0.213 & 0.214 & 0.250 & 0.270 & 0.328 & 0.399 & 0.466 & 0.542 & 0.622 & 0.704 & 0.234 \\ \hline
  	0.5 & 0.263 & 0.242 & 0.281 & 0.287 & 0.326 & 0.391 & 0.468 & 0.545 & 0.626 & 0.703 & 0.271 \\ \hline
  	0.6 & 0.317 & 0.276 & 0.342 & 0.323 & 0.342 & 0.393 & 0.463 & 0.546 & 0.623 & 0.698 & 0.321 \\ \hline
  	0.7 & 0.372 & 0.326 & 0.398 & 0.362 & 0.365 & 0.399 & 0.456 & 0.528 & 0.611 & 0.691 & 0.372 \\ \hline
  	\end{tabular}}
  	\caption{RMSE obtained by the Park and Kim SVQR model with different $\sigma$ and $\epsilon$ values for $\tau$=0.9. }
  	\label{table33}
  \end{table}

	\begin{table}[]
		\centering
		\begin{tabular}{|l|l|l|l|l|l|l|l|}
			\hline
			$\sigma$             & 0.100 & 0.200 & 0.300 & 0.400 & 0.500 & 0.600 & 0.700 \\ \hline
			SVQR              & 0.059 & 0.112 & 0.166 & 0.213 & 0.263 & 0.317 & 0.372 \\ \hline
		   Park and Kim SVQR & 0.059 & 0.112 & 0.166 & 0.213 & 0.242 & 0.276 & 0.326 \\ \hline
			Sparse SVQR       & 0.059 & 0.112 & 0.166 & 0.213 & 0.258 & 0.294 & 0.353 \\ \hline
			$\epsilon$-SVQR     & 0.046 & 0.090 & 0.136 & 0.182 & 0.227 & 0.276 & 0.335 \\ \hline
		\end{tabular}
	\caption{Minimum RMSE obtained by proposed  different SVQR models with different values of $\sigma$ for $\tau=0.9$.}
	\label{table4}
	\end{table}

     \begin{table}
     \begin{tabular}{|c|c|c|c|c|}
     \hline
        & RMSE             & MAE              & TheilU           & (C,s,$\epsilon$) \\ \hline
                   \multicolumn{5}{|l|}{ $\tau$=0.1 }       \\ \hline
     SVQR        & 0.2002    0.0166 & 0.1609    0.0152 & 0.2244    0.0224 & (4,1,0)                       \\ \hline
     Sparse SVQR & 0.2052    0.0195 & 0.1683    0.0160 & 0.2247    0.0244 & (4,1,0.1)                     \\ \hline
      $\epsilon$-SVQR   & 0.1483    0.0091 & 0.1242    0.0075 & 0.1337    0.0087 & (4,1,5)                       \\ \hline
          \multicolumn{5}{|l|}{ $\tau$=0.3 }       \\ \hline
     SVQR        & 0.3545    0.0113 & 0.2869    0.0159 & 0.3304    0.0132 & (4,1,0)                       \\ \hline
     Sparse SVQR & 0.2881    0.0283 & 0.2400    0.0215 & 0.2787    0.0286 & (4,1,0.4)                     \\ \hline
      $\epsilon$-SVQR   & 0.2641    0.0147 & 0.2181    0.0115 & 0.2482    0.0129 & (4,1,1.5)                     \\ \hline
    \multicolumn{5}{|l|}{ $\tau$=0.7 }       \\ \hline
     SVQR        & 0.7864    0.0180 & 0.6912    0.0130 & 0.7665    0.0239 & (4,1,0)                       \\ \hline
     Sparse SVQR & 0.7051    0.0128 & 0.6149    0.0100 & 0.6827    0.0207 & (4,1,0.4)                     \\ \hline
      $\epsilon$-SVQR   & 0.6556    0.0344 & 0.5345    0.0173 & 0.6289    0.0358 & (4,1,1.3)                     \\ \hline
    \multicolumn{5}{|l|}{ $\tau$=0.9 }       \\ \hline
     SVQR        & 1.3056    0.0953 & 1.1127    0.0604 & 1.2542    0.0967 & (4,1,0)                       \\ \hline
     Sparse SVQR & 0.9470    0.0278 & 0.8176    0.0204 & 0.9016    0.0326 & (4,1,0.4)                     \\ \hline
      $\epsilon$-SVQR   & 0.9222    0.0422 & 0.8123    0.0260 & 0.8855    0.0445 & (4,1,4)                       \\ \hline
     \end{tabular}
     \caption{Cmparision of SVQR , Sparse SVQR and proposed $\epsilon$-SVQR model on AD2 artificial datasets}
     \label{table5}
     \end{table}
    
            We have also compared the performance of the Sparse SVQR model with the proposed $\epsilon$-SVQR model on artificial dataset AD2 with different evaluation criteria. The AD2 artifical dataset contains asymmetric noise from $\chi^2(3)$.  Table \ref{table5} shows the optimal performance of the standard SVQR , Sparse SVQR model and proposed $\epsilon$-SVQR model using different evaluation criteria for different $\tau$ values along with tunned values of parameters. It can be observed that the Sparse SVQR and proposed $\epsilon$-SVQR model obtains better generalization ability than the standard SVQR model. It means that the use of $\epsilon$-insensitive zone in SVQR model helps it to own better generalization ability. Further, it can also be observed that the proposed $\epsilon$-SVQR model owns better generalization ability than Sparse SVQR model.   
     
     \subsection{UCI Datasets}
     We have also considered UCI datasets to show the efficacy of the proposed $\epsilon$-SVQR model. For this, we have downloaded the Servo (~167$\times$5~) , Boston Housing (~506$\times$14~) and Traizines (~186$\times$61~) datasets from the UCI repository \cite{UCIbenchmark}. The  $80 \%$ of the data points were used for the training the regression model where as remaining $20\%$ of data points were used for the testing regression estimate. Since the actual true quantile functions for these UCI datasets are unknown, so we have computed the Error $E_{\tau}$ by computing the converge probability described in Section \ref{Perform_criteria} in 100 trails for evaluation of performances of SVQR models. It is noteworthy that the standard SVQR model is equivalent to the proposed  $\epsilon$-SVQR model with $\epsilon$ ~=~0. 
     
\begin{figure}
   	\centering
   	\includegraphics[width = 6.0in,height=4.0in]{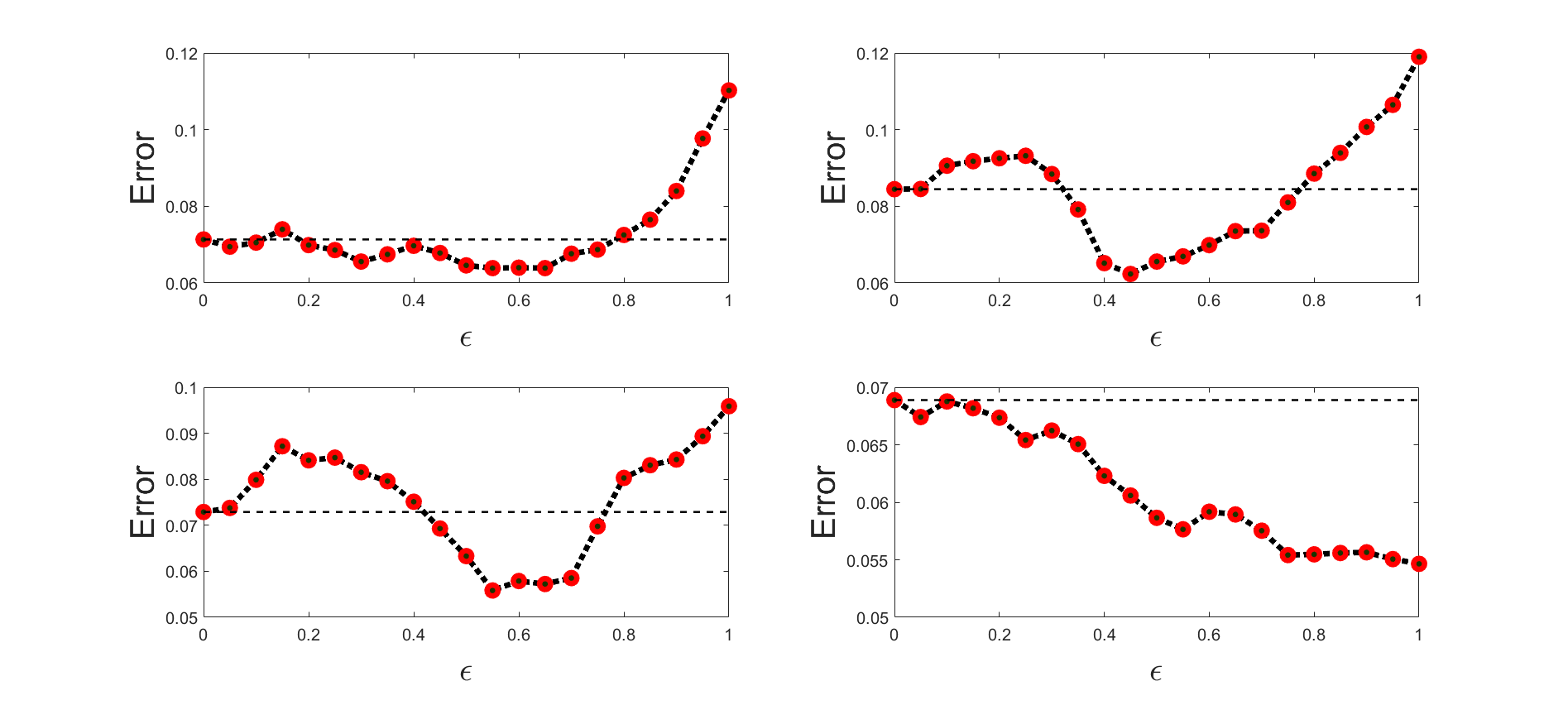}
   	\caption{Error obtained by proposed $\epsilon$-SVQR model on Servo dataset with different $\epsilon$ values for $\tau = 0.4$ , $\tau=0.6$ ,$\tau=0.7$ and $\tau=0.8$ respectively.  }
   	\label{fig:servoeps}
   \end{figure}

\begin{figure}
   	\centering
   	\includegraphics[width = 4.5in,height=2.5in]{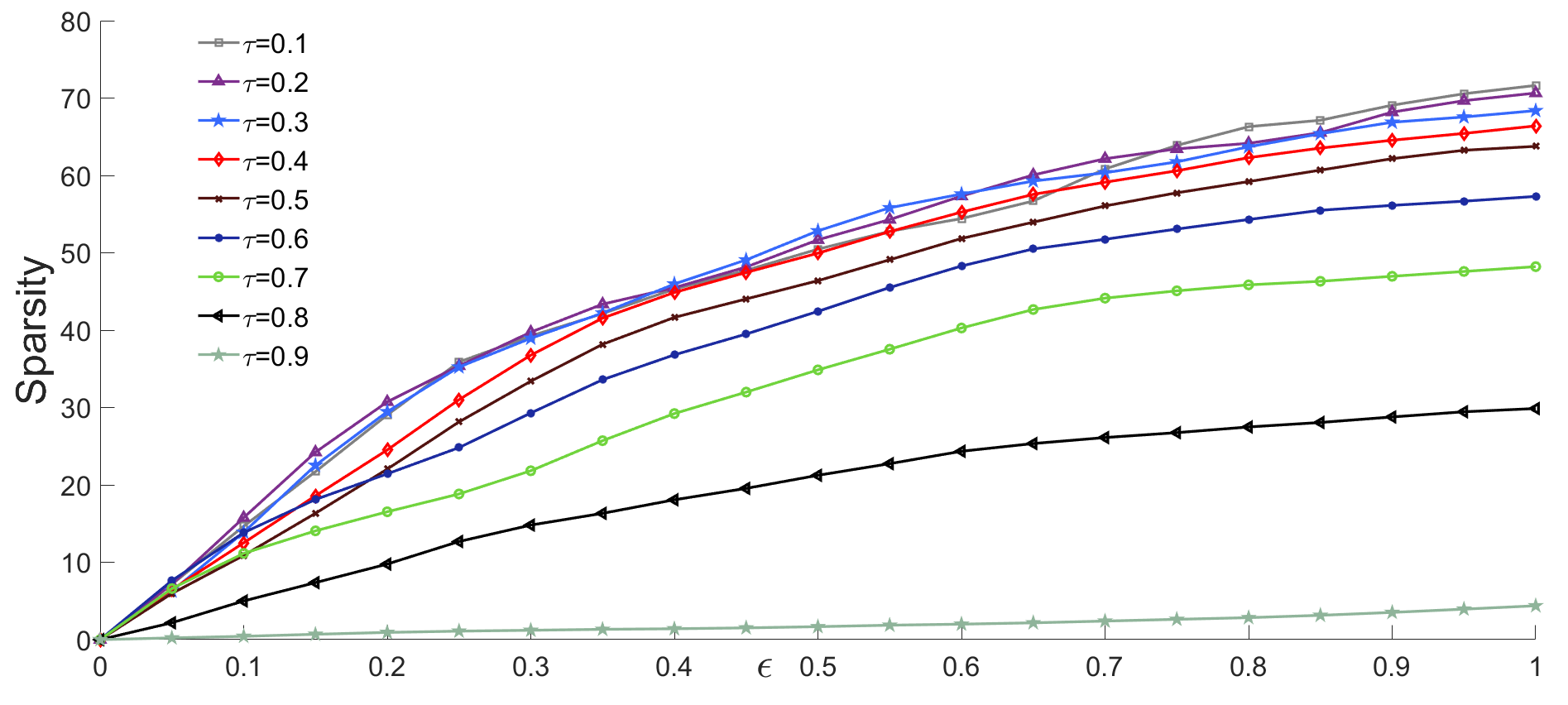}
   	\caption{Sparsity obtained by proposed $\epsilon$-SVQR model on Servo dataset with different $\epsilon$ values for different $\tau$ values.}
   	\label{servo_esvqr_spar}
   \end{figure}

\begin{figure}
   	\centering
   	\includegraphics[width=0.8\linewidth]{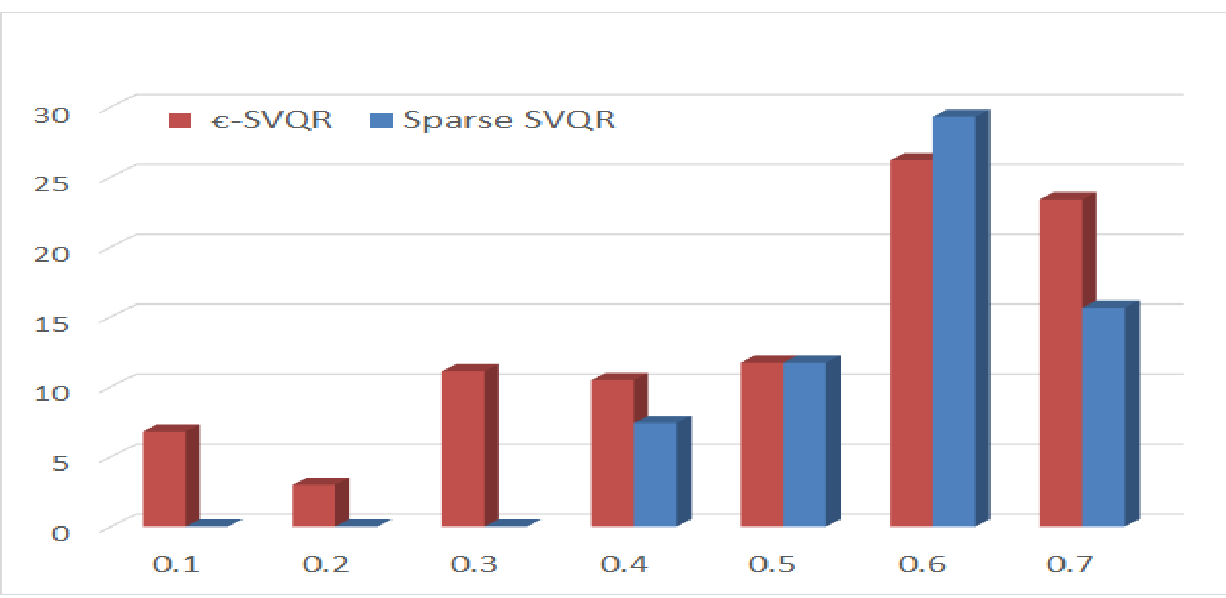}
   	\caption{ Percentage of the improvement in RMSE obtained by proposed $\epsilon$-SVQR and Sparse SVQR model over standard SVQR model for different value of $\tau$. }
   	\label{servo11}
   \end{figure}

         Table \ref{servo_esvqr} lists the Error $E_{\tau}$ obtained by the proposed $\epsilon$-SVQR model with different $\epsilon$ values for different $\tau$ values on the Servo dataset.  It can be observed that there exists several choices of non-zero values of $\epsilon$ for which, the proposed $\epsilon$-SVQR can outperform the existing standard SVQR model. It is well depicted by the  Figure (\ref{fig:servoeps}). It can be realized that the proposed  $\epsilon$-SVQR model is a better substitute of SVQR model. Figure (\ref{servo_esvqr_spar})  shows the plot of sparsity obtained by $\epsilon$-SVQR model against different $\epsilon$ values for different quantiles. It can be observed that irrespective of values of $\epsilon$, the proposed $\epsilon$-SVQR model can obtain sparse solution which increases with the increase in $\epsilon$ value. Table \ref{servo_spasevqr} lists the Error $E_{\tau}$ obtained by the Sparse SVQR model with different $\epsilon$ values for different $\tau$ values on the Servo dataset. The Sparse SVQR model can  also obtain the improvement over SVQR model by tunning the $\epsilon$ values but, it is only on few $\tau$ values. For $\tau$ =  0.1,0.2,0.3,  the Sparse SVQR model hardly improves the SVQR model. The possible reason behind this fact is that the Sparse SVQR model fails to control the effective width of the $\epsilon$-insensitive zone for lower values of $\tau$. Figure (\ref{servo11}) compares the Sparse SVQR model and proposed $\epsilon$-SVQR model in the terms of percentage of improvement in Error obtained by these models over SVQR model. It can be observed that the proposed $\epsilon$-SVQR is always a better substitute than Sparse SVQR model.

          Table \ref{boston-hosuing} and \ref{traizines} list the performance of the proposed $\epsilon$-SVQR model with the different $\epsilon$ values in the estimation of different quantiles for Boston Housing (~506$\times$14~) and Traizines (~186$\times$61~) datasets respectively . The similar kinds of observation can also be drawn from these tables.
   	\begin{table}[]
   	\centering
   	{\fontsize{10}{10} \selectfont
   		\begin{tabular}{|l|l|l|l|l|l|l|l|l|}
   			\hline
   			$\epsilon$/$\tau$ & 0.1   & 0.2   & 0.3   & 0.4   & 0.5   & 0.6   & 0.7   & 0.8   \\ \hline
   			0        & 0.040 & 0.055 & 0.067 & 0.071 & 0.078 & 0.084 & 0.073 & 0.069 \\ \hline
   			0.05     & 0.041 & 0.053 & 0.065 & 0.069 & 0.081 & 0.085 & 0.074 & 0.067 \\ \hline
   			0.10      & 0.037 & 0.054 & 0.064 & 0.071 & 0.081 & 0.091 & 0.080 & 0.069 \\ \hline
   			0.15     & 0.038 & 0.056 & 0.060 & 0.074 & 0.080 & 0.092 & 0.087 & 0.068 \\ \hline
   			0.20      & 0.039 & 0.056 & 0.059 & 0.070 & 0.083 & 0.093 & 0.084 & 0.067 \\ \hline
   			0.25     & 0.037 & 0.056 & 0.060 & 0.069 & 0.083 & 0.093 & 0.085 & 0.065 \\ \hline
   			0.30      & 0.039 & 0.069 & 0.059 & 0.066 & 0.081 & 0.088 & 0.082 & 0.066 \\ \hline
   			0.35     & 0.042 & 0.064 & 0.062 & 0.067 & 0.079 & 0.079 & 0.080 & 0.065 \\ \hline
   			0.40      & 0.045 & 0.060 & 0.065 & 0.070 & 0.076 & 0.065 & 0.075 & 0.062 \\ \hline
   			0.45     & 0.044 & 0.054 & 0.064 & 0.068 & 0.075 & 0.062 & 0.069 & 0.061 \\ \hline
   			0.50      & 0.043 & 0.054 & 0.064 & 0.065 & 0.072 & 0.066 & 0.063 & 0.059 \\ \hline
   			0.55     & 0.044 & 0.059 & 0.069 & 0.064 & 0.069 & 0.067 & 0.056 & 0.058 \\ \hline
   			0.60      & 0.043 & 0.064 & 0.072 & 0.064 & 0.069 & 0.070 & 0.058 & 0.059 \\ \hline
   			0.65     & 0.043 & 0.071 & 0.071 & 0.064 & 0.071 & 0.074 & 0.057 & 0.059 \\ \hline
   			0.70     & 0.044 & 0.074 & 0.067 & 0.068 & 0.069 & 0.074 & 0.059 & 0.058 \\ \hline
   			0.75     & 0.044 & 0.072 & 0.061 & 0.069 & 0.077 & 0.081 & 0.070 & 0.055 \\ \hline
   			0.8      & 0.043 & 0.070 & 0.062 & 0.073 & 0.093 & 0.089 & 0.080 & 0.055 \\ \hline
   			0.85     & 0.040 & 0.072 & 0.068 & 0.077 & 0.103 & 0.094 & 0.083 & 0.056 \\ \hline
   			0.90      & 0.040 & 0.072 & 0.073 & 0.084 & 0.109 & 0.101 & 0.084 & 0.056 \\ \hline
   			0.95     & 0.041 & 0.076 & 0.073 & 0.098 & 0.119 & 0.106 & 0.089 & 0.055 \\ \hline
   			1.00        & 0.040 & 0.074 & 0.073 & 0.110 & 0.131 & 0.119 & 0.096 & 0.055 \\ \hline
   	\end{tabular}}
   	\caption{Error obtained by the proposed $\epsilon$-SVQR model with different $\epsilon$ values for different $\tau$ values on Servo dataset }
   	\label{servo_esvqr}
   \end{table}
   
   
   \begin{table}[]
   	\centering
   	{\fontsize{10}{10} \selectfont
   		\begin{tabular}{|l|l|l|l|l|l|l|l|l|}
   			\hline
   			$\epsilon$/$\tau$ & 0.1   & 0.2   & 0.3   & 0.4   & 0.5   & 0.6   & 0.7   & 0.8   \\ \hline
   			0        & 0.040 & 0.055 & 0.067 & 0.071 & 0.078 & 0.084 & 0.073 & 0.069 \\ \hline
   			0.05     & 0.051 & 0.069 & 0.070 & 0.077 & 0.081 & 0.087 & 0.073 & 0.059 \\ \hline
   			0.1      & 0.056 & 0.089 & 0.076 & 0.079 & 0.083 & 0.083 & 0.063 & 0.057 \\ \hline
   			0.15     & 0.058 & 0.112 & 0.101 & 0.078 & 0.081 & 0.071 & 0.061 & 0.056 \\ \hline
   			0.2      & 0.059 & 0.115 & 0.124 & 0.081 & 0.076 & 0.060 & 0.065 & 0.055 \\ \hline
   			0.25     & 0.057 & 0.119 & 0.129 & 0.084 & 0.072 & 0.077 & 0.083 & 0.055 \\ \hline
   			0.3      & 0.055 & 0.120 & 0.145 & 0.073 & 0.069 & 0.090 & 0.096 & 0.055 \\ \hline
   			0.35     & 0.051 & 0.118 & 0.147 & 0.075 & 0.069 & 0.101 & 0.108 & 0.056 \\ \hline
   			0.4      & 0.051 & 0.117 & 0.142 & 0.071 & 0.093 & 0.122 & 0.111 & 0.058 \\ \hline
   			0.45     & 0.052 & 0.114 & 0.133 & 0.066 & 0.109 & 0.151 & 0.115 & 0.056 \\ \hline
   			0.5      & 0.090 & 0.111 & 0.115 & 0.067 & 0.131 & 0.168 & 0.117 & 0.056 \\ \hline
   			0.55     & 0.095 & 0.104 & 0.096 & 0.076 & 0.153 & 0.178 & 0.118 & 0.056 \\ \hline
   			0.6      & 0.100 & 0.098 & 0.092 & 0.094 & 0.176 & 0.191 & 0.118 & 0.058 \\ \hline
   			0.65     & 0.100 & 0.092 & 0.077 & 0.114 & 0.196 & 0.196 & 0.118 & 0.060 \\ \hline
   			0.7      & 0.100 & 0.090 & 0.070 & 0.129 & 0.204 & 0.199 & 0.119 & 0.069 \\ \hline
   			0.75     & 0.100 & 0.087 & 0.067 & 0.158 & 0.214 & 0.201 & 0.121 & 0.071 \\ \hline
   			0.8      & 0.100 & 0.081 & 0.072 & 0.180 & 0.226 & 0.206 & 0.123 & 0.071 \\ \hline
   			0.85     & 0.100 & 0.075 & 0.075 & 0.200 & 0.242 & 0.209 & 0.128 & 0.072 \\ \hline
   			0.9      & 0.100 & 0.071 & 0.086 & 0.217 & 0.252 & 0.214 & 0.129 & 0.076 \\ \hline
   			0.95     & 0.100 & 0.070 & 0.110 & 0.230 & 0.258 & 0.214 & 0.129 & 0.079 \\ \hline
   			1        & 0.100 & 0.074 & 0.137 & 0.244 & 0.265 & 0.215 & 0.129 & 0.081 \\ \hline
   	\end{tabular}}
   	\caption{Error obtained by the proposed Sparse SVQR model with different $\epsilon$ for different $\tau$ values on Servo dataset }
   	\label{servo_spasevqr}
   \end{table}
   
 \begin{table}[]
	\centering
	{\fontsize{10}{10} \selectfont
		\begin{tabular}{|l|l|l|l|l|l|l|l|l|}
			\hline
			$\epsilon$/$\tau$ & 0.1   & 0.2   & 0.3   & 0.4   & 0.5   & 0.6   & 0.7   & 0.8   \\ \hline
		0        & 0.040 & 0.055 & 0.067 & 0.071 & 0.078 & 0.084 & 0.073 & 0.069 \\ \hline
		0.05     & 0.043 & 0.068 & 0.060 & 0.070 & 0.083 & 0.093 & 0.085 & 0.067 \\ \hline
		0.1      & 0.038 & 0.067 & 0.065 & 0.071 & 0.076 & 0.064 & 0.065 & 0.059 \\ \hline
		0.15     & 0.036 & 0.076 & 0.064 & 0.062 & 0.069 & 0.071 & 0.060 & 0.055 \\ \hline
		0.2      & 0.044 & 0.060 & 0.072 & 0.076 & 0.093 & 0.093 & 0.090 & 0.055 \\ \hline
		0.25     & 0.068 & 0.060 & 0.091 & 0.117 & 0.131 & 0.128 & 0.109 & 0.055 \\ \hline
		0.3      & 0.116 & 0.120 & 0.160 & 0.173 & 0.176 & 0.161 & 0.112 & 0.058 \\ \hline
		0.35     & 0.196 & 0.203 & 0.212 & 0.219 & 0.204 & 0.185 & 0.116 & 0.056 \\ \hline
		0.4      & 0.055 & 0.289 & 0.258 & 0.252 & 0.226 & 0.198 & 0.118 & 0.056 \\ \hline
		0.45     & 0.068 & 0.327 & 0.296 & 0.287 & 0.252 & 0.201 & 0.118 & 0.056 \\ \hline
		0.5      & 0.095 & 0.352 & 0.330 & 0.303 & 0.265 & 0.206 & 0.119 & 0.056 \\ \hline
		0.55     & 0.100 & 0.386 & 0.374 & 0.334 & 0.275 & 0.212 & 0.123 & 0.056 \\ \hline
		0.6      & 0.100 & 0.413 & 0.396 & 0.344 & 0.290 & 0.215 & 0.129 & 0.056 \\ \hline
		0.65     & 0.100 & 0.438 & 0.418 & 0.361 & 0.298 & 0.216 & 0.129 & 0.056 \\ \hline
		0.7      & 0.100 & 0.389 & 0.437 & 0.372 & 0.303 & 0.216 & 0.129 & 0.056 \\ \hline
		0.75     & 0.100 & 0.314 & 0.447 & 0.384 & 0.309 & 0.216 & 0.129 & 0.056 \\ \hline
		0.8      & 0.100 & 0.207 & 0.457 & 0.396 & 0.312 & 0.216 & 0.129 & 0.056 \\ \hline
		0.85     & 0.100 & 0.064 & 0.461 & 0.396 & 0.316 & 0.219 & 0.129 & 0.056 \\ \hline
		0.9      & 0.100 & 0.070 & 0.461 & 0.397 & 0.316 & 0.220 & 0.129 & 0.056 \\ \hline
		0.95     & 0.100 & 0.108 & 0.442 & 0.400 & 0.316 & 0.223 & 0.129 & 0.056 \\ \hline
		1        & 0.100 & 0.140 & 0.420 & 0.402 & 0.316 & 0.221 & 0.129 & 0.056 \\ \hline
	\end{tabular}}
	\caption{Error obtained by the proposed Park and Kim SVQR model with different $\epsilon$ for different $\tau$ values on Servo dataset }
	\label{servo_spasevqr}
\end{table}


	\begin{table}[]
		\centering
		\begin{tabular}{|l|l|l|l|l|l|l|l|l|}
			\hline
			$\tau$               & 0.1   & 0.2   & 0.3   & 0.4   & 0.5   & 0.6   & 0.7   & 0.8   \\ \hline
			SVQR              & 0.040 & 0.055 & 0.067 & 0.071 & 0.078 & 0.084 & 0.073 & 0.069 \\ \hline
			$\epsilon$-SVQR            & 0.037 & 0.053 & 0.059 & 0.064 & 0.069 & 0.062 & 0.056 & 0.055 \\ \hline
			Sparse SVQR       & 0.040 & 0.055 & 0.067 & 0.066 & 0.069 & 0.060 & 0.061 & 0.055 \\ \hline
			Park and Kim SVQR & 0.036 & 0.055 & 0.060 & 0.062 & 0.069 & 0.064 & 0.060 & 0.055 \\ \hline
		\end{tabular}
	\caption{Minimum RMSE obtained by different SVQR model for different $\tau$ values.}
	\end{table}



%

   \begin{table}[]
   \begin{tabular}{|l|l|l|l|l|}
   \hline
              &                 & Error            & Sparsity & CPU time (s) \\ \hline
              & $\epsilon$=0    & 0.0287$~\pm~$0.0223 & 0.00     & 12.44        \\ \cline{2-5}
   $\tau$=0.1 & $\epsilon$=0.8  & 0.0223$~\pm~$0.0196 & 2.20     & 13.65        \\ \cline{2-5}
              & $\epsilon$=0.9  & 0.0225$~\pm~$0.0190 & 2.51     & 13.39        \\ \cline{2-5}
              & $\epsilon$=1    & 0.0226$~\pm~$0.0185 & 2.89     & 13.93     \\ \hline
    & $\epsilon$=0    & 0.0447$~\pm~$0.0363 & 0.00     & 13.02        \\ \cline{2-5}
       $\tau$=0.5       & $\epsilon$= 0.5 & 0.0435$~\pm~$0.0360 & 5.79     & 13.25        \\ \cline{2-5}
              & $\epsilon$ =0.6 & 0.0433$~\pm~$0.0359 & 7.14     & 13.94        \\ \cline{2-5}
              & $\epsilon$= 0.7 & 0.0436$~\pm~$0.0366 & 4.54     & 13.10        \\ \hline
    & $\epsilon$=0    & 0.0393$~\pm~$0.0278 & 0.00     & 14.60        \\ \cline{2-5}
              & $\epsilon$=3    & 0.0315$~\pm~$0.0227 & 12.34    & 14.08        \\ \cline{2-5}
       $\tau$=0.9       & $\epsilon$= 4   & 0.0317$~\pm~$0.0209 & 16.84    & 14.42        \\ \cline{2-5}
              & $\epsilon$=5    & 0.0286$~\pm~$0.0198 & 21.17    & 15.07    \\ \hline
   \end{tabular}
        	\caption{Performance of the proposed $\epsilon$-SVQR model on Boston Housing dataset for different value of $\tau$}
        	\label{boston-hosuing}
   \end{table}
   
   \begin{table}[]
   \begin{tabular}{|l|l|l|l|l|}
   \hline
              &                & Error            & Sparsity & CPU time (s) \\ \hline
              & $\epsilon$=0   & 0.0481$~\pm~$0.0430  & 0.00     & 2.54         \\\cline{2-5}
   $\tau$=0.1 & $\epsilon$=0.2 & 0.0450$~\pm~$0.0351 & 29.55    & 2.63         \\ \cline{2-5}
              & $\epsilon$=0.3 & 0.0433$~\pm~$ 0.0280 & 48.64    & 2.55         \\ \hline

   $\tau$=0.3 & $\epsilon$=0   & 0.0722$~\pm~$0.0518 & 0.00     & 2.59         \\ \cline{2-5}
              & $\epsilon$=0.1 & 0.0679$~\pm~$0.0473 & 27.68    & 2.51         \\ \hline
   & $\epsilon$=0   & 0.0699$~\pm~$0.0602 & 0.00     & 2.49         \\ \cline{2-5}
       $\tau$=0.7        & $\epsilon$=0.1 & 0.0623$~\pm~$0.0479 & 39.41    & 2.54         \\ \cline{2-5}
              & $\epsilon$=0.2 & 0.0598$~\pm~$0.0494 & 62.38    & 2.43         \\ \hline
        $\tau$=0.8       & $\epsilon$=0   & 0.0549$~\pm~$0.0459 & 0.00     & 2.56         \\ \cline{2-5}
   & $\epsilon$=0.1 & 0.0537$~\pm~$0.0407 & 33.61    & 2.57         \\ \hline
   \end{tabular}
           	\caption{Performance of the proposed $\epsilon$-SVQR model on Traizines dataset for different value of $\tau$}
           	\label{traizines}
   \end{table}

   \section{Conclusion}
    This paper reviews the development of Support Vector Quantile Regression (SVQR) models on the line of popular $\epsilon$- Support Vector Regression model and finds that the existing  pinball loss functions fail to incorporate the $\epsilon$-insensitive zone in true sense.  Further, this paper proposes a novel asymmetric $\epsilon$-insensitive pinball loss function for measuring the empirical risk in support vector quantile regression model. The proposed  asymmetric $\epsilon$-insensitive pinball loss function divides the fixed width of $\epsilon$-insensitive zone using the $\tau$ value and present a suitable asymmetric $\epsilon$-insensitive zone for each $\tau$ value. The resulting model which has been termed with '`$\epsilon$-Support Vector Quantile Regression` ($\epsilon$-SVQR) model  ignores data points which lie inside of the asymmetric $\epsilon$-insensitive zone which make the proposed model a sparse regression model. The optimal choice of the value of $\epsilon$ in the proposed $\epsilon$-SVQR model depends upon the variance present in the training data points around the $\epsilon$-insensitive zone. The $\epsilon$-SVQR model improves the prediction of the existing SVQR model significantly and also enjoys the sparsity as well. The detailed experiments carried on various artifical and real world UCI datasets using the various evaluation creteria show that the proposed $\epsilon$-SVQR owns better generalization ability than other SVQR models.

%
%
%
%
%
   
                      The choice of the value of $\epsilon$ is crucial in the proposed $\epsilon$-SVQR model. Therefore, we would like to develop techniques to obtain the optimal choice of the $\epsilon$ for a given training set. 
     
     \section*{Acknowledgment}
          We would like to acknowledge Ministry of Electronics and Information Technology, Government of India, as this work has been funded by them under Visvesvaraya PhD Scheme for Electronics and IT, Order No. Phd-MLA/4(42)/2015-16. 
      \section* {Conflict of Interest}
       We authors hereby declare that we do not have any conflict of interest with the content of this manuscript.

\end{document}